\documentclass[runningheads]{llncs}
\usepackage{graphicx}
\usepackage{amsmath,amssymb} 
\usepackage{color}
\usepackage[width=122mm,left=12mm,paperwidth=146mm,height=193mm,top=12mm,paperheight=217mm]{geometry}
\usepackage{subcaption}
\usepackage[linesnumbered,ruled]{algorithm2e}
\usepackage{booktabs}
\usepackage{hyperref}

\newcommand{\overbar}[1]{\mkern 1.5mu\overline{\mkern-1.5mu#1\mkern-1.5mu}\mkern 1.5mu}

\newcommand{\eg}{\emph{e.g.,}\xspace}
\newcommand{\etal}{\emph{et al.}\xspace}
\newcommand{\ie}{\emph{i.e.,}\xspace}

\DeclareMathOperator*{\argmax}{arg\,max}

\begin{document}
\pagestyle{headings}
\mainmatter

\title{Spot On: Action Localization from Pointly-Supervised Proposals}


\authorrunning{Pascal Mettes \and Jan C. van Gemert \and Cees G. M. Snoek}

\author{Pascal Mettes$^{\ast}$ \and Jan C. van Gemert$^{\ddagger}$ \and Cees G. M. Snoek$^{\ast}$}
\institute{$^{\ast}$University of Amsterdam\\$^{\ddagger}$Delft University of Technology}

\maketitle

\begin{abstract}
We strive for spatio-temporal localization of actions in videos. The state-of-the-art relies on action proposals at test time and selects the best one with a classifier trained on carefully annotated box annotations. Annotating action boxes in video is cumbersome, tedious, and error prone. Rather than annotating boxes, we propose to annotate actions in video with points on a sparse subset of frames only. We introduce an overlap measure between action proposals and points and incorporate them all into the objective of a non-convex Multiple Instance Learning optimization. Experimental evaluation on the UCF Sports and UCF 101 datasets shows that 
(i) spatio-temporal proposals can be used to train classifiers while retaining the localization performance, 
(ii) point annotations yield results comparable to box annotations while being significantly faster to annotate,
(iii) with a minimum amount of supervision our approach is competitive to the state-of-the-art.
Finally, we introduce spatio-temporal action annotations on the train and test videos of Hollywood2, resulting in \emph{Hollywood2Tubes}, available at \texttt{\url{http://tinyurl.com/hollywood2tubes}}.
 \keywords{Action localization, action proposals}
\end{abstract}

\section{Introduction}
This paper is about spatio-temporal localization of actions like \emph{Driving a car}, \emph{Kissing}, and \emph{Hugging} in videos. Starting from a sliding window legacy \cite{TianPartCVPR2013}, the common approach these days is to generate tube-like proposals at test time, encode each of them with a feature embedding and select the most relevant one, \eg \cite{jain2014action,yuCVPR2015fap,vangemert2015apt,soomroICCV2015actionLocContextWalk}. All these works, be it sliding windows or tube proposals, assume that a carefully annotated training set with boxes per frame is available a priori. In this paper, we challenge this assumption. We propose a simple algorithm that leverages proposals at \emph{training} time, with a minimum amount of supervision, to speedup action location annotation.

\begin{figure}[t]
\centering
\includegraphics[width=\textwidth]{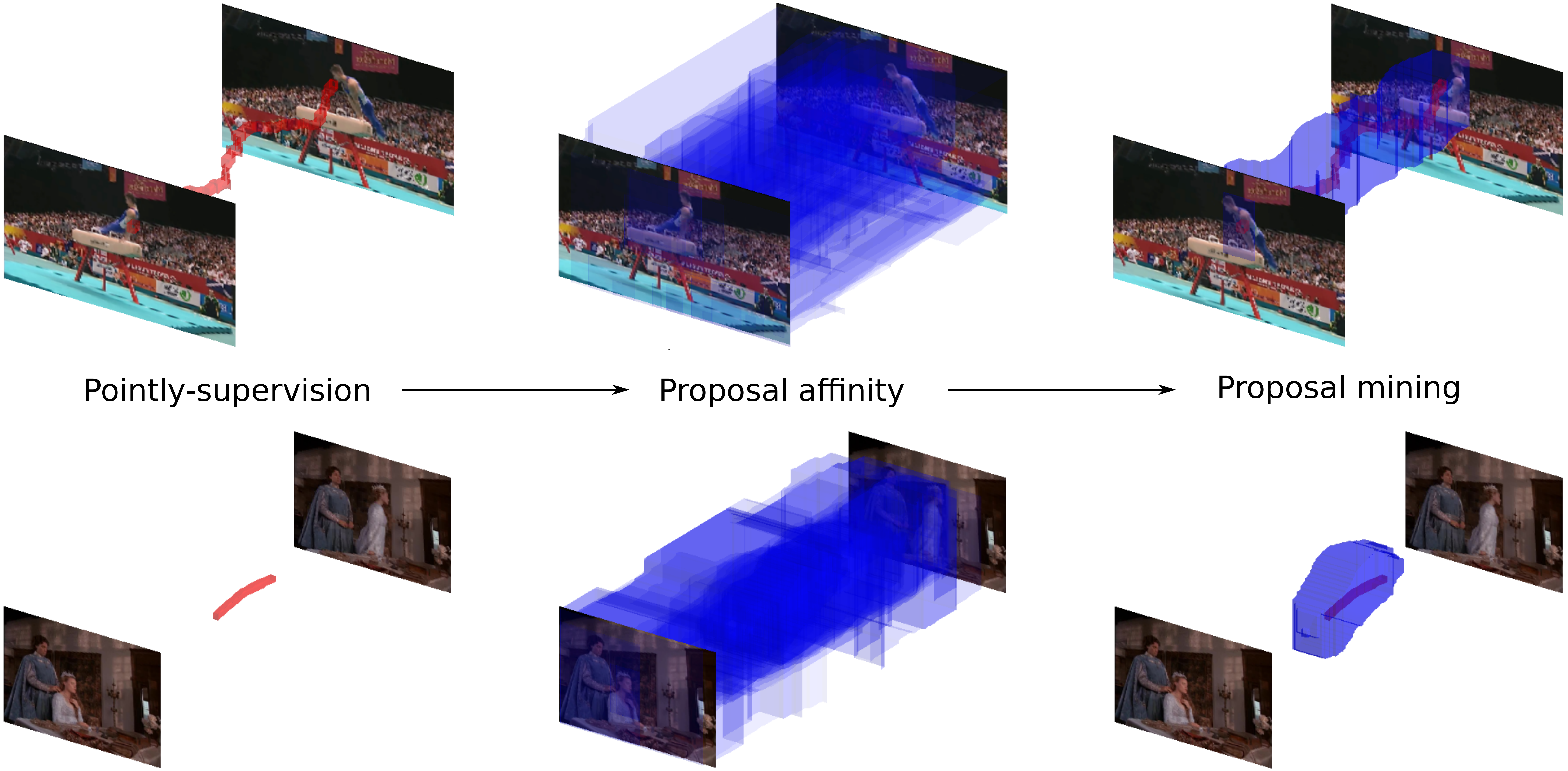}
\caption{\textbf{Overview of our approach} for a \emph{Swinging} and \emph{Standing up} action. First, the video is annotated cheaply using point-supervision. Then, action proposals are extracted and scored using our overlap measure. Finally, our proposal mining aims to discover the single one proposal that best represents the action, given the provided points.}
\label{fig:qual-method}
\end{figure}

We draw inspiration from related work on weakly-supervised object detection, \eg \cite{KimNIPS2009,RussakovskyECCV2012,cinbis2014multi}. The goal is to detect an object and its bounding box at test time given only the object class label at train time and no additional supervision. The common tactic in the literature is to model this as a Multiple Instance Learning (MIL) problem \cite{cinbis2014multi,NguyenICCV2009,andrews2002support} where positive images contain at least one positive object proposal and negative images contain only negative proposals. During each iteration of MIL, a detector is trained and applied on the train set to re-identify the object proposal most likely to enclose the object of interest. Upon convergence, the final detector is applied on the test set. Methods typically vary in their choice of initial proposals and the multiple instance learning optimization. 
In the domain of action localization a similar MIL tactic easily extends to action proposals as well but results in poor accuracy as our experiments show. Similar to weakly-supervised object detection, we rely on (action) proposals and MIL, but we include a minimum amount of supervision to retain action localization accuracy competitive with full supervision.

Obvious candidates for the supervision are action class labels and bounding boxes, but other forms of supervision, such as tags and line strokes, are also feasible~\cite{XuCVPR2015}. In~\cite{bearmanArXiv15whatsthepoint}, Bearman \etal show that human-provided points on the image are valuable annotations for semantic segmentation of objects. By inclusion of an objectness prior in their loss function they report a better efficiency/effectiveness trade off compared to image-level annotations and free-from squiggles.
We follow their example in the video domain and leverage point-supervision to aid MIL in finding the best action proposals at training time.

We make three contributions in this work. First, we propose to train action localization classifiers using spatio-temporal proposals as positive examples rather than ground truth tubes. While common in object detection, such an approach is as of yet unconventional in action localization. In fact, we show that using proposals instead of ground truth annotations does not lead to a decrease in action localization accuracy. Second, we introduce an MIL algorithm that is able to mine proposals with a good spatio-temporal fit to actions of interest by including point supervision. It extends the traditional MIL objective with an overlap measure that takes into account the affinity between proposals and points. Finally, with the aid of our proposal mining algorithm, we are able to supplement the complete Hollywood2 dataset by Marsza{\l}ek \etal \cite{marszalek09} with action location annotations, resulting in \emph{Hollywood2Tubes}. We summarize our approach in Figure~\ref{fig:qual-method}. Experiments on Hollywood2Tubes, as well as the more traditional UCF Sports and UCF 101 collections support our claims. Before detailing our pointly-supervised approach we present related work.

\section{Related work}
\label{sec:relwork}




Action localization is a difficult problem and annotations are avidly used. Single image bounding box annotations allow training a part-based detector~\cite{TianPartCVPR2013,lan2011discriminative} or a per-frame detector where results are aggregated over time~\cite{gkioxari2015finding,weinzaepfelICCV2015learningToTrack}. However, since such detectors first have to be trained themselves, they cannot be used when no bounding box annotations are available. Independent training data can be brought in to automatically detect individual persons for action localization~\cite{yuCVPR2015fap,lucorsoCVPR2015humanAction,wang2014video}. 
A person detector, however, will fail to localize contextual actions such as \textit{Driving} or interactions such as \textit{Shaking hands} or \textit{Kissing}. Recent work using unsupervised action proposals based on supervoxels~\cite{jain2014action,soomroICCV2015actionLocContextWalk,oneata2014spatio} or on trajectory clustering
\cite{vangemert2015apt,chencorsoICCV2015actiondetectionMotionClustering,marianICCV2015unsupervisedTube}, have shown good results for action localization. In this paper we rely on action proposals to aid annotation. Proposals give excellent recall without supervision and are thus well-suited for an unlabeled train set.







Large annotated datasets are slowly coming available in action localization. Open annotations benefit the community, paving the way for new data-driven action localization methods. UCF-Sports~\cite{soomro2014actionInSports}, HOHA~\cite{raptis2012discovering} and MSR-II~\cite{cao2010crossDatasetActionDetectionMSRIIset} have up to a few hundred actions, while UCF101~\cite{soomro2012ucf101}, Penn-Action~\cite{zhangICCV13actemes}, and J-HMBD~\cite{jhuangICCV2013towardsUnderstanding} have 1--3 thousand action clips and 3 to 24 action classes. The problem of scaling up to larger sets is not due to sheer dataset size: there are millions of action videos with hundreds of action classes available~\cite{soomro2012ucf101,gorban2015thumos,karpathy2014largescalevidSports1M,kuehne2011hmdb}. The problem lies with the spatio-temporal annotation effort. 
In this paper we show how to ease this annotation effort, exemplified by releasing spatio-temporal annotations for all Hollywood2 \cite{marszalek09} videos.

Several software tools are developed to lighten the annotation burden. The gain can come from a well-designed user interface to annotate videos with bounding boxes~\cite{mihalcik2003ViPER,vondrickIJCV2013crowdsourced} or even polygons~\cite{yuenICCV09labelmeVideo}. We move away from such complex annotations and only require a point. Such point annotations can readily be included in existing annotation tools which would further reduce effort. Other algorithms can reduce annotation effort by intelligently selecting which example to label~\cite{settles2010active}. Active  learning~\cite{vondrick2011video} or trained detectors~\cite{biancoCVIU15interactiveAnnotation} can assist the human annotator. The disadvantage of such methods is the bias towards the used recognition method. We do not bias any algorithm to decide where and what to annotate: by only setting points we can quickly annotate all videos.



Weakly supervised methods predict more information than was annotated. Examples from static images include predicting a bounding box while having only class labels \cite{cinbis2014multi,bilenCVPR15weakObjDetConvexClust,OquabCVPR15isObjLocForFree} or even no labels al all~\cite{choCVPR15unsupervised}. In the video domain, the temporal dimension offers more annotation variation. Semi-supervised learning for video object detection is done with a few bounding boxes~\cite{aliCVPR11flowboost,MisraCVPR15semiSupObjeDetfromVid}, a few global frame labels~\cite{wangECCV14videoObject}, only video class labels~\cite{sivaECCV12defenceNegativeMining}, or no labels at all~\cite{kwakICCV15unsupervisedObjectInVid}. For action localization, only the video label is used by~\cite{mosabbeb2014multi,siva2011weakly}, whereas \cite{jain2015objects2action} use no labels. As our experiments show, using no label or just class labels performs well below fully supervised results. Thus, we propose a middle ground: pointing at the action. Compared to annotating full bounding boxes this greatly reduces annotation time while retaining accuracy.




\section{Strong action localization using cheap annotations}
We start from the hypothesis that an action localization proposal may substitute the ground truth on a training set without a significant loss of classification accuracy. Proposal algorithms yield hundreds to thousands of proposals per video with the hope that at least one proposal matches the action well~\cite{jain2014action,vangemert2015apt,soomroICCV2015actionLocContextWalk,oneata2014spatio,chencorsoICCV2015actiondetectionMotionClustering,marianICCV2015unsupervisedTube}. The problem thus becomes how to mine the best proposal out of a large set of candidate proposals with minimal supervision effort.

\subsection{Cheap annotations: action class labels and pointly-supervision}
A minimum of supervision effort is an action class label for the whole video. For such global video labels, a traditional approach to mining the best proposal is Multiple Instance Learning~\cite{andrews2002support} (MIL). In the context of action localization, each video is interpreted as a bag and the proposals in each video are interpreted as its instances. The goal of MIL is to train a classifier that can be used for proposal mining by using only the global label. 

Next to the global action class label we leverage cheap annotations within each video: for a subset of frames we simply point at the action. We refer to such a set of point annotations as \textit{pointly-supervision}. The supervision allows us to easily exclude those proposals that have no overlap with any annotated point. Nevertheless, there are still many proposals that intersect with at least one point. Thus, points do not uniquely identify a single proposal. In the following we will introduce an overlap measure to associate proposals with points. To perform the proposal mining, we will extend MIL's objective to include this measure.

\subsection{Measuring overlap between points and proposals}
To explain how we obtain our overlap measure, let us first introduce the following notation. For a video $V$ of $N$ frames, an action localization proposal $A=\{\text{BB}_i  \}_{\text{i}=f}^m$ consists of connected bounding boxes through video frames $(f,...,m)$ where $1 \le f \le m \le N$. We use $\overbar{BB_{i}}$ to indicate the center of a bounding box $i$. The pointly-supervision $C=\{(x_i,y_i) \}^K$ is a set of $K \le N$ sub-sampled video frames where each frame $i$ has a single annotated point $(x_i,y_i)$. Our overlap measure outputs a score for each proposal depending on how well the proposal matches the points. 

Inspired by a mild center-bias in annotators~\cite{tseng2009quantifying}, we introduce a term $M(\cdot)$ to represent how close the center of a bounding box proposal is to an annotated point, relative to the bounding box size. Since large proposals have a higher likelihood to contain any annotated point we use a regularization term $S(\cdot)$ on the proposal size. The center-bias term $M(\cdot)$ normalizes the distance to the bounding box center by the distance to the furthest bounding box side. A point $(x_i,y_i) \in C$ outside a bounding box $BB_i \in A$ scores 0 and a point on the bounding box center $\overbar{BB_{i}}$ scores 1. The score decreases linearly with the distance to the center for the point. It is averaged over all annotated points $K$:
\begin{equation}
M(A, C)  =   \frac{1}{K} \sum_{i=1}^{K}  \text{max}(0, 1 - \frac{||(x_i,y_i) - \overbar{BB_{K_i}} ||_2}{ \max_{(u,v) \in e(BB_{K_i})} ||( (u,v) - \overbar{BB_{K_i}}) ||_2},
\label{eq:overlap1}
\end{equation}
where $e(BB_{K_i})$ denotes the box edges of box $BB_{K_i}$.

We furthermore add a regularization on the size of the proposals. 
The idea behind the regularization is that small spatial proposals can occur anywhere. Large proposals, however, are obstructed by the edges of the video. This biases their middle-point around the center of the video, where the action often happens. 
The size regularization term $S(\cdot)$ addresses this bias by penalizing proposals with large bounding boxes $|BB_{i}| \in A$, compared to the size of a video frame $|F_i| \in V$,
\begin{equation}
S(A, V)  =  \big( \frac{ \sum_{i=f}^m |BB_{i}| }{\sum_{j=1}^N |F_j|} \big) ^{2}.
\end{equation}

Using the center-bias term $M(\cdot)$ regularized by $S(\cdot)$, our overlap measure $O(\cdot)$ is defined as 
\begin{equation}
O(A, C, V)  =  M(A, C) - S(A, V).
\label{eq:overlappoint}
\end{equation}
Recall that $A$ are the proposals, $C$ captures the pointly-supervision and $V$ the video. We use $O(\cdot)$ in an iterative proposal mining algorithm over all annotated videos in search for the best proposals.

\subsection{Mining proposals overlapping with points}
For proposal mining, we start from a set of action videos $\{ \textbf{x}_{i}, t_{i}, y_{i}, C_i\}_{i=1}^{N}$, where $\textbf{x}_{i} \in \mathbb{R}^{A_{i} \times D}$ is the $D$-dimensional feature representation of the $A_{i}$ proposals in video $i$. Variable $t_{i} = \{ \{ BB_{j} \}_{j=f}^{m} \}^{A_{i}}$ denotes the collection of tubes for the $A_{i}$ proposals. Cheap annotations consist of the class label $y_i$ and the points $C_i$.

For proposal mining we insert our overlap measure $O(\cdot)$ in a Multiple Instance Learning scheme to train a classification model that can learn the difference between good and bad proposals. Guided by $O(\cdot)$, the classifier becomes increasingly more aware about which proposals are a good representative for an action. We start from a standard MIL-SVM~\cite{cinbis2014multi,andrews2002support} and adapt it's objective with the the mining score $P(\cdot)$ of each proposal, which incorporates our function $O(\cdot)$ as:

\begin{equation}
\begin{split}
 & \min_{\mathbf{w},b,\xi} \frac{1}{2} ||\mathbf{w}||^{2} + \lambda \sum_{i} \xi_{i},\\
\text{s.t.} \quad & \forall_{i} : y_{i} \cdot ( \mathbf{w} \cdot \argmax_{\mathbf{z} \in x_{i}} P(\mathbf{z} | \mathbf{w}, b, t_{i}, C_{i}, V_i) + b) \geq 1 - \xi_{i},\\
&\forall_{i} : \xi_i \geq 0,
\end{split}
\label{eq:milsvm}
\end{equation}
where $(\mathbf{w},b)$ denote the classifier parameters, $\xi_{i}$ denotes the slack variable and $\lambda$ denotes the regularization parameter. The proposal with the highest mining score per video is used to train the classifier. 

The objective of Equation~\ref{eq:milsvm} is non-convex due to the joint minimization over the classifier parameters $(\mathbf{w}, b)$ and the maximization over the mined proposals $P(\cdot)$. Therefore, we perform iterative block coordinate descent by alternating between clamping one and optimizing the other. For fixed classifier parameters $(\mathbf{w}, b)$, we mine the proposal with the highest Maximum a Posteriori estimate with the classifier as the likelihood and $O(\cdot)$ as the prior:
\begin{eqnarray}
P(\mathbf{z} | \mathbf{w}, b, t_{i}, C_{i}, V_i) & \propto & \left( <\!\!\mathbf{w}, \mathbf{z}\!\!> + b \right) \cdot O(t_i, C_{i}, V_i).
\label{eq:mining-map}
\end{eqnarray}  
After a proposal mining step, we fix $P(\cdot)$ and train the classifier parameters $(\mathbf{w}, b)$ with stochastic gradient descent on the mined proposals. We alternate the mining and classifier optimizations for a fixed amount of iterations.  After the iterative optimization, we train a final SVM on the best mined proposals and use that classifier for action localization.

\section{Experimental setup}
\label{sec:experiments}

\subsection{Datasets}

\noindent
We perform our evaluation on two action localization datasets that have bounding box annotations both for training and test videos.
\\\\
\textbf{UCF Sports} consists of 150 videos covering 10 action categories \cite{RodriguezCVPR2008}, such as \emph{Diving}, \emph{Kicking}, and \emph{Skateboarding}. The videos are extracted from sport broadcasts and are trimmed to contain a single action. We employ the train and test data split as suggested in~\cite{lan2011discriminative}.
\\
\textbf{UCF 101} has 101 actions categories \cite{soomro2012ucf101} where 24 categories have spatio-temporal action localization annotations. This subset has 3,204 videos, where each video contains a single action category, but might contain multiple instances of the same action. We use the first split of the train and test sets as suggested in~\cite{soomro2012ucf101} with 2,290 videos for training and 914 videos for testing.

\subsection{Implementation details}
\label{sec:details}

\indent
\textbf{Proposals.} Our proposal mining is agnostic to the underlying proposal algorithm. We have performed experiments using proposals from both APT~\cite{vangemert2015apt} and Tubelets~\cite{jain2014action}. We found APT to perform slightly better and report all results using APT.
\\
\textbf{Features.} For each tube we extract Improved Dense Trajectories and compute HOG, HOF, Traj, MBH features~\cite{wang13}. The combined features are reduced to 128 dimensions through PCA and aggregated into a fixed-size representation using Fisher Vectors~\cite{sanchez2013image}.
We construct a codebook of 128 clusters, resulting in a 54,656-dimensional representation per proposal.
\\
\textbf{Training.} We train the proposal mining optimization for 10 iterations for all our evaluations, similar to Cinbis \etal~\cite{cinbis2014multi}. Following further suggestions by~\cite{cinbis2014multi}, we randomly split the training videos into multiple (3) splits to train and select the instances. While training a classifier for one action, we randomly sample 100 proposals of each video from the other actions as negatives. We set the SVM regularization $\lambda$ to 100.
\\
\textbf{Evaluation.} During testing we apply the classifier to all proposals of a test video and maintain the top proposals per video. To evaluate the action localization performance, we compute the Intersection-over-Union (IoU) between proposal $p$ and the box annotations of the corresponding test example $b$ as: $\text{iou}(p, b) = \frac{1}{| \Gamma |} \sum_{f \in \Gamma} IoU_{p,b}(f)$, where $\Gamma$ is the set of frames where at least one of $p,b$ is present~\cite{jain2014action}. The function $IoU$ states the box overlap for a specified frame.
For IoU threshold $t$, a top selected proposal is deemed a positive detection if $\text{iou}(p, b) \geq t$.

After combining the top proposals from all videos, we compute the Average Precision score using their ranked scores and positive/negative detections. For the comparison to the state-of-the-art on UCF Sports, we additionally report AUC (Area under ROC curve) on the scores and detections.

\section{Results}

\begin{figure}[t]
\centering
\begin{subfigure}{0.49\textwidth}
\centering
\includegraphics[width=\textwidth]{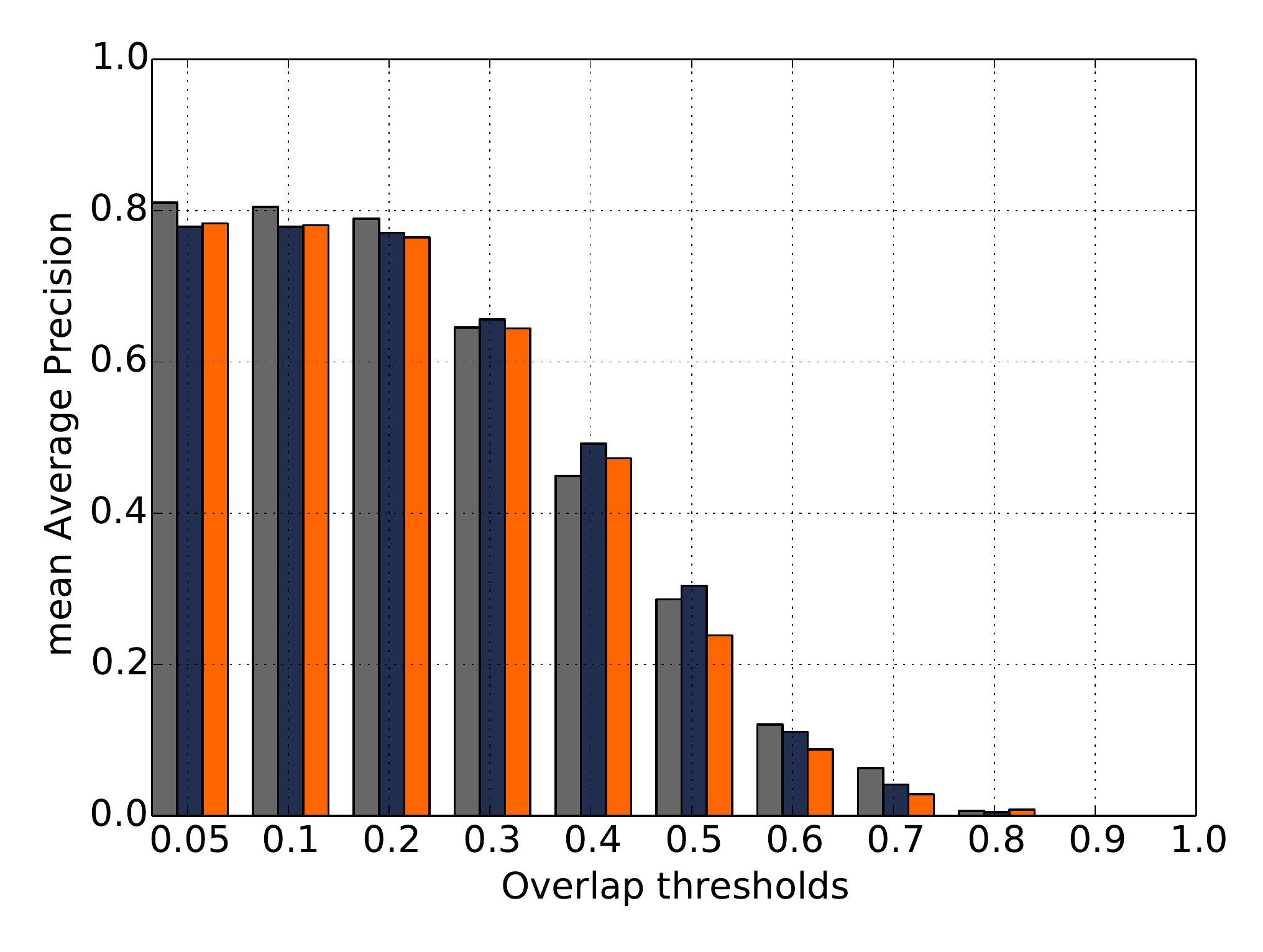}
\caption{UCF Sports.}
\end{subfigure}
\begin{subfigure}{0.49\textwidth}
\centering
\includegraphics[width=\textwidth]{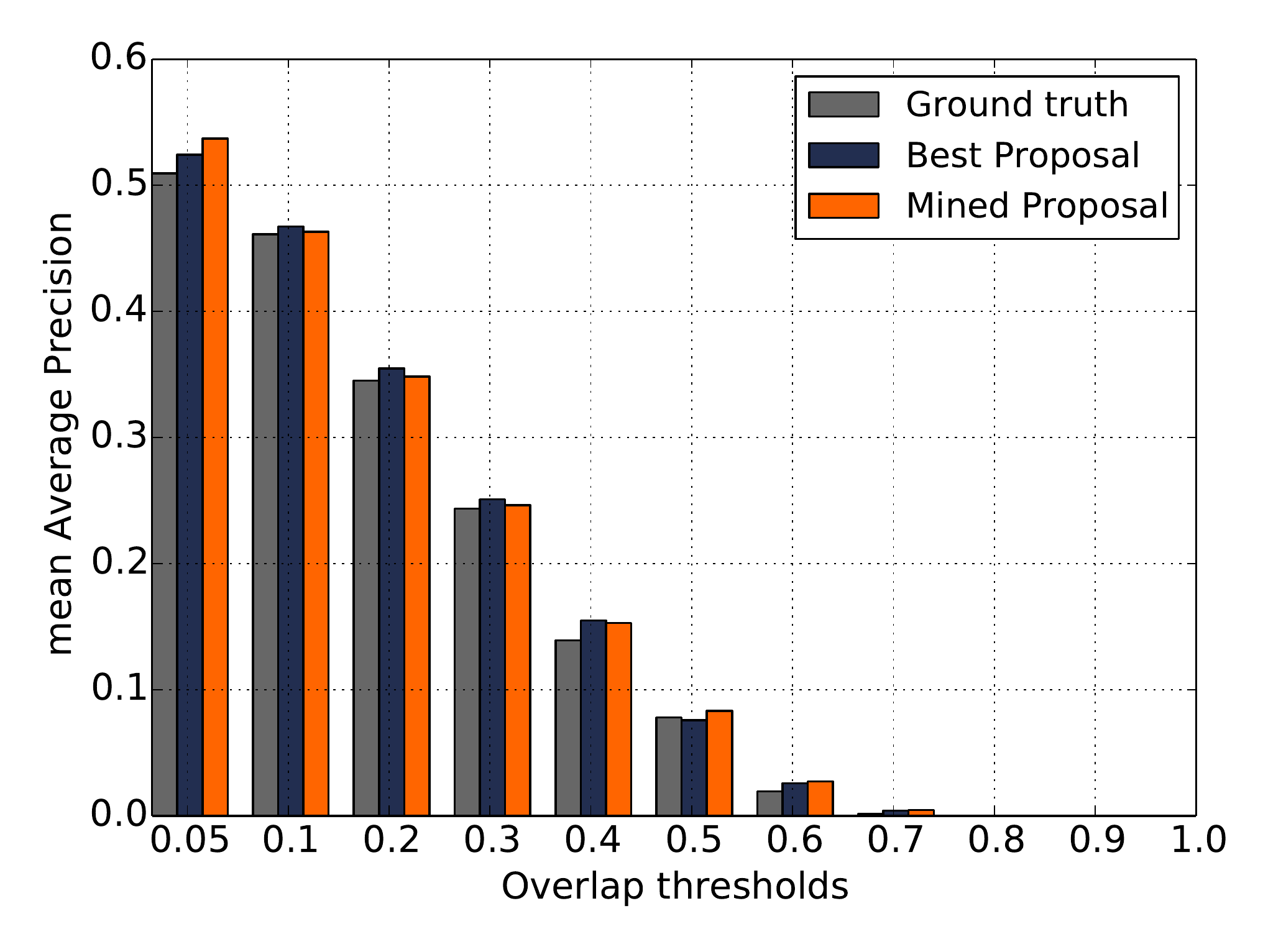}
\caption{UCF 101.}
\end{subfigure}
\caption{\textbf{Training action localization classifiers with proposals} vs ground truth tubes on (a) UCF Sports and (b) UCF 101. Across both datasets and thresholds, the best possible proposal yields similar results to using the ground truth. Also note how well our mined proposal matches the ground truth and best possible proposal we could have selected.}
\label{fig:exp1}
\end{figure}

\subsection{Training without ground truth tubes}
First we evaluate our starting hypothesis of replacing ground truth tubes with proposals for training action localization classifiers. We compare three approaches: 1) train on ground truth annotated bounding boxes; 2) train on the proposal with the highest IoU overlap for each video; 3) train on the proposal mined based on point annotations and our proposal mining. For the points on both datasets, we take the center of each annotated bounding box.

\textbf{Training with the best proposal.} Figure~\ref{fig:exp1} shows that the localization results for the best proposal are similar to the ground truth tube for both datasets and across all IoU overlap thresholds as defined in Section~\ref{sec:details}.
This result shows that proposals are sufficient to train classifiers for action localization. The result is somewhat surprising given that the best proposals used to train the classifiers have a less than perfect fit with the ground truth action. We computed the fit with the ground truth, and on average the IoU score of the best proposals (the ABO score) is 0.642 on UCF Sports and 0.400 on UCF 101. The best proposals are quite loosely aligned with the ground truth. Yet, training on such non-perfect proposals is not detrimental for results. This means that a perfect fit with the action is not a necessity during training. An explanation for this result is that the action classifier is now trained on the same type of noisy samples that it will encounter at test-time. This better aligns the training with the testing, resulting in slightly improved accuracy.

\begin{figure}[t]
\centering
\begin{subfigure}{0.3\textwidth}
\includegraphics[width=\textwidth]{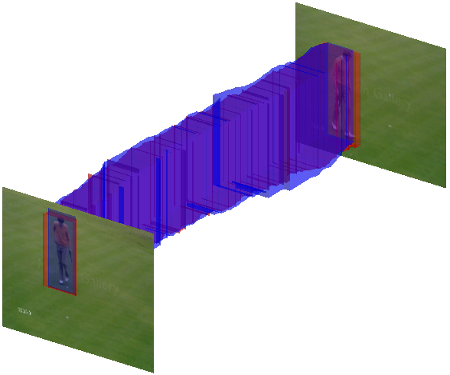}
\caption{\emph{Walking.}}
\label{fig:exp1-qual-1}
\end{subfigure}
\hspace{0.25cm}
\begin{subfigure}{0.3\textwidth}
\includegraphics[width=\textwidth]{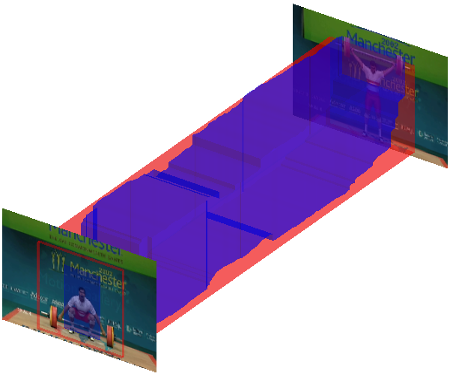}
\caption{\emph{Lifting.}}
\label{fig:exp1-qual-2}
\end{subfigure}
\hspace{0.25cm}
\begin{subfigure}{0.3\textwidth}
\includegraphics[width=\textwidth]{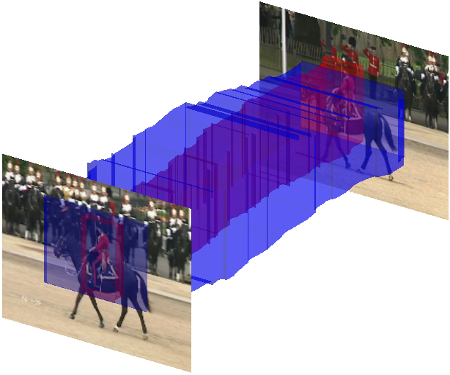}
\caption{\emph{Riding horse.}}
\label{fig:exp1-qual-3}
\end{subfigure}
\caption{\textbf{Training video showing our mined proposal} (blue) and the ground truth (red). (a) Mined proposals might have a high similarity to the ground truth. In (b) our mining focuses solely on the person lifting, while in (c) our mining has learned to include part of the horse. An imperfect fit with the ground truth does not imply a bad proposal.}
\label{fig:exp1-qual}
\end{figure}

\textbf{Training with proposal mining from points.} Figure~\ref{fig:exp1} furthermore shows the localization results from training without bounding box annotations using only point annotations. On both data sets, results are competitive to the ground truth tubes across all thresholds. This result shows that when training on proposals, carefully annotated box annotations are not required. 
Our proposal mining is able to discover the best proposals from cheap point annotations. The discrepancy between the ground truth and our mined proposal for training is shown in Figure~\ref{fig:exp1-qual} for thee videos. For some videos, \eg Figure~\ref{fig:exp1-qual-1}, the ground truth and the proposal have a high similarity. This does however not hold for all videos, \eg Figures~\ref{fig:exp1-qual-2}, where our mined proposal focuses solely on the lifter (\emph{Lifting}), and~\ref{fig:exp1-qual-3}, where our mined proposal includes the horse (\emph{Horse riding}).

\textbf{Analysis.} On UCF 101, where actions are not temporally trimmed, we observe an average temporal overlap of 0.74. The spatial overlap in frames where proposals and ground truth match is 0.38. This result indicates that we are better capable of detecting actions in the temporal domain than the spatial domain. On average, top ranked proposals during testing are 2.67 times larger than their corresponding ground truth. Despite a preference for larger proposals, our results are comparable to the fully supervised method trained on expensive ground truth bounding box tubes. Finally, we observe that most false positives are proposals from positive test videos with an overlap score below the specified threshold. On average, 26.7\% of the top 10 proposals on UCF 101 are proposals below the overlap threshold of 0.2. Regarding false negatives, on UCF 101 at a 0.2 overlap threshold, 37.2\% of the actions are not among the top selected proposals. This is primarily because the proposal algorithm does not provide a single proposal with enough overlap.

From this experiment we conclude that training directly on proposals does not lead to a reduction in action localization accuracy. Furthermore, using cheap point annotations with our proposal mining yields results competitive to using carefully annotated bounding box annotations.

\subsection{Must go faster: lowering the annotation frame-rate}

\begin{figure}[t]
\centering
\begin{subfigure}{\textwidth}
\centering
\includegraphics[width=0.479\textwidth]{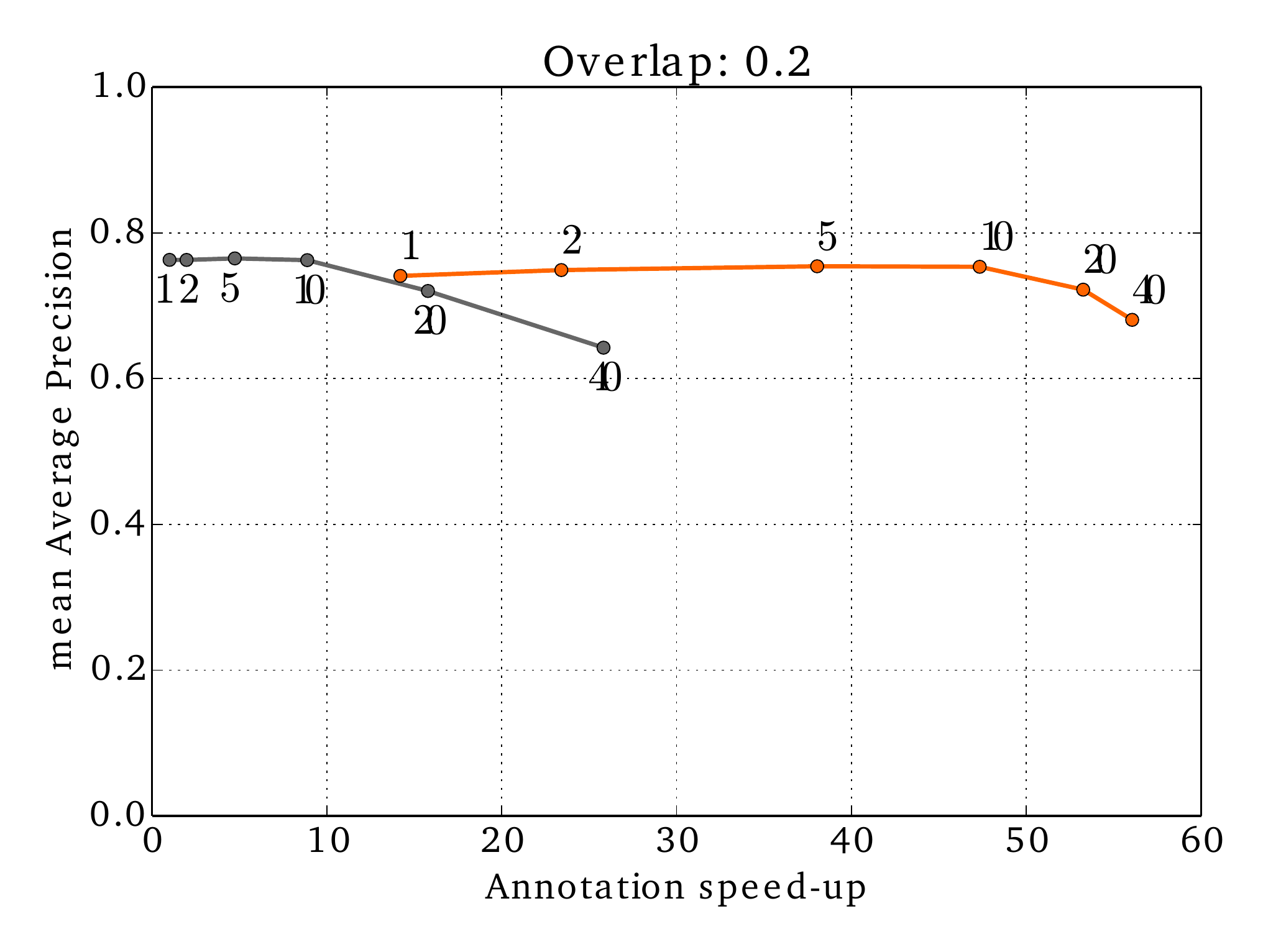}
\includegraphics[width=0.479\textwidth]{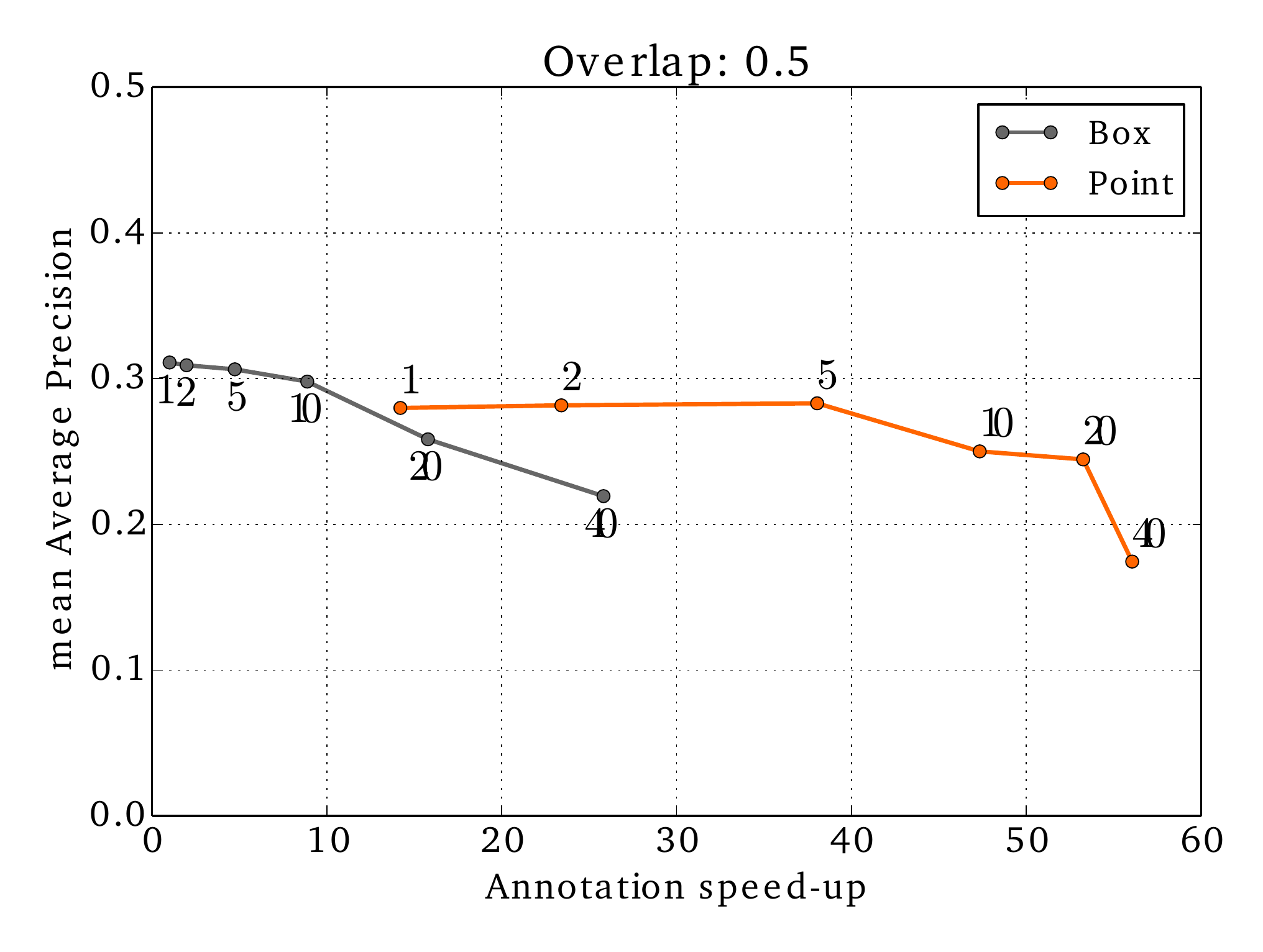}
\caption{UCF Sports.}
\label{fig:exp2-sports}
\end{subfigure}
\begin{subfigure}{\textwidth}
\centering
\includegraphics[width=0.479\textwidth]{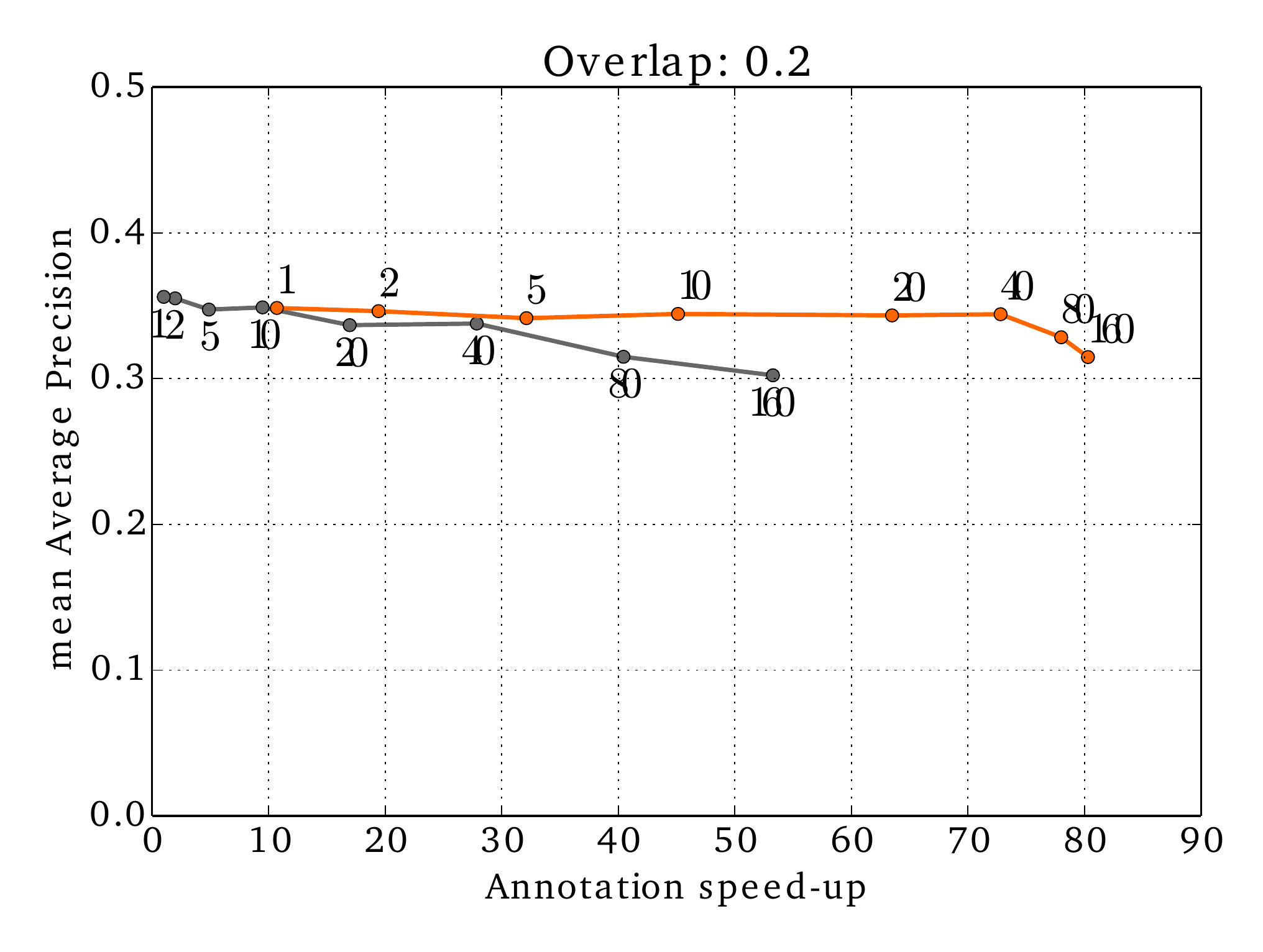}
\includegraphics[width=0.479\textwidth]{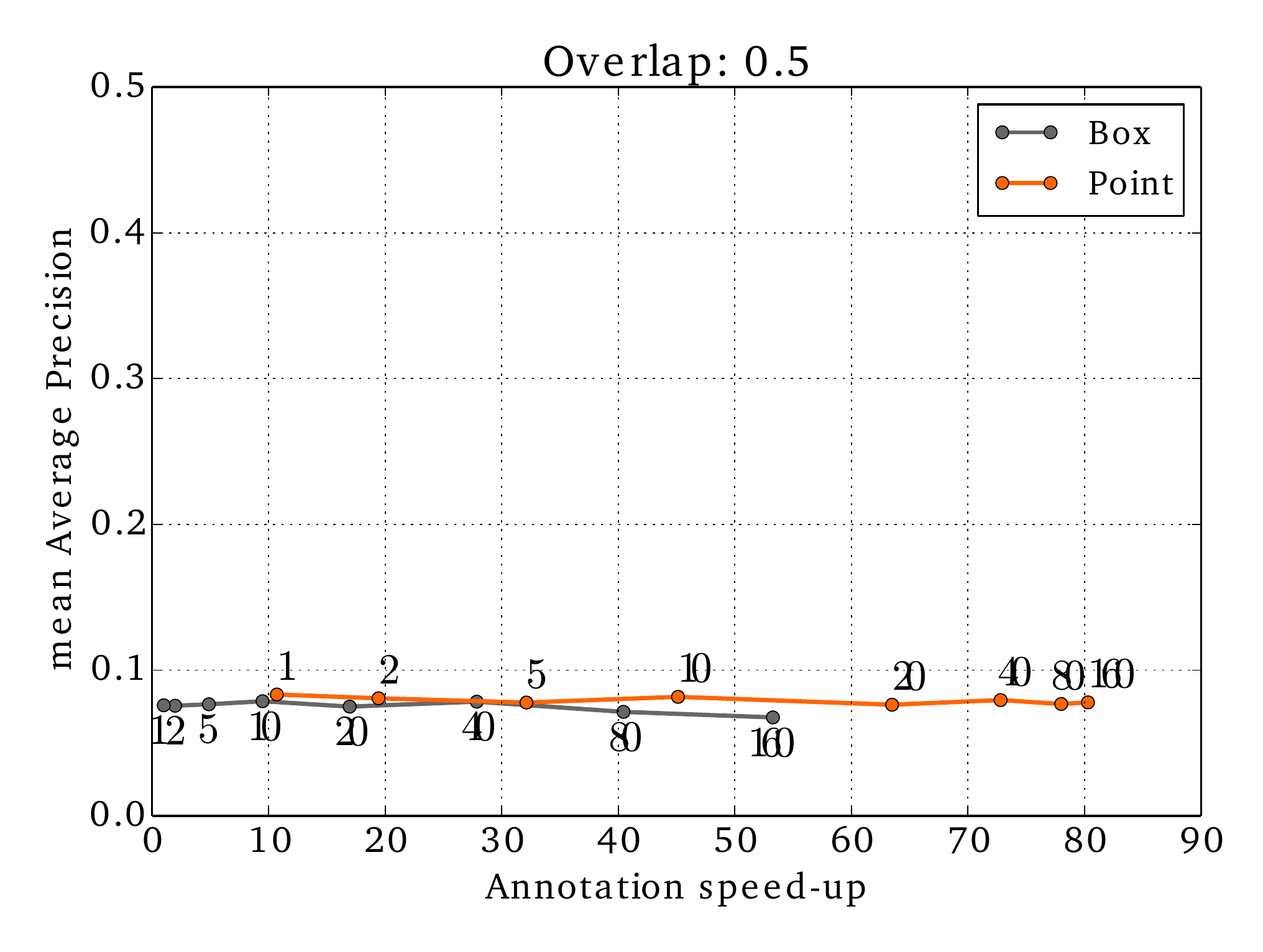}
\caption{UCF 101.}
\label{fig:exp2-101}
\end{subfigure}
\caption{\textbf{The annotation speedup} versus mean Average Precision scores on (a) UCF Sports and (b) UCF 101 for two overlap thresholds using both box and point annotations. The annotation frame-rates are indicated on the lines. Using points remains competitive to boxes with a 10x to 80x annotation speed-up.}
\label{fig:exp2}
\end{figure}

The annotation effort can be significantly reduced by annotating less frames. Here we investigate how a higher annotation frame-rate influences the trade-off between annotation speed-up versus classification performance. We compare higher annotation frame-rates for points and ground-truth bounding boxes. 

\textbf{Setup.} For measuring annotation time we randomly selected 100 videos from the UCF Sports and UCF 101 datasets separately and performed the annotations. We manually annotated boxes and points for all evaluated frame-rates $\{1,2,5,10,...\}$. We obtain the points by simply reducing a bounding box annotation to its center. We report the speed-up in annotation time compared to drawing a bounding box on every frame.  
 Classification results are given for two common IoU overlap thresholds on the test set, namely 0.2 and 0.5. 

\textbf{Results.} In Figure~\ref{fig:exp2} we show the localization performance as a function of the annotation speed-up for UCF Sports and UCF 101. Note that when annotating all frames, a point is roughly 10-15 times faster to annotate than a box. The reason for the reduction in relative speed-up between the higher frame-rates is due to the constant time spent on determining the action label of each video. When analyzing classification performance we note it is not required to annotate all frames. Although the performance generally decreases as less frames are annotated, using a frame rate of 10 (\ie annotating 10\% of the frames) is generally sufficient for retaining localization performance. We can get competitive classification scores with an annotation speedup of 45 times or more.

\begin{figure}[t]
\centering
\begin{subfigure}{0.875\textwidth}
\centering
\includegraphics[width=\textwidth]{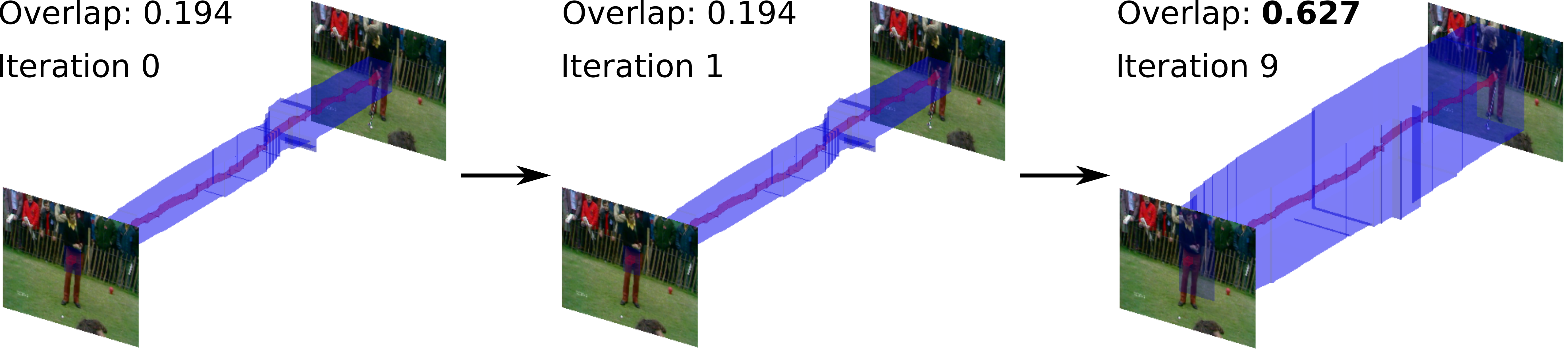}
\caption{\emph{Swinging Golf.}}
\end{subfigure}
\vspace{0.05cm}\\
\begin{subfigure}{0.875\textwidth}
\centering
\includegraphics[width=\textwidth]{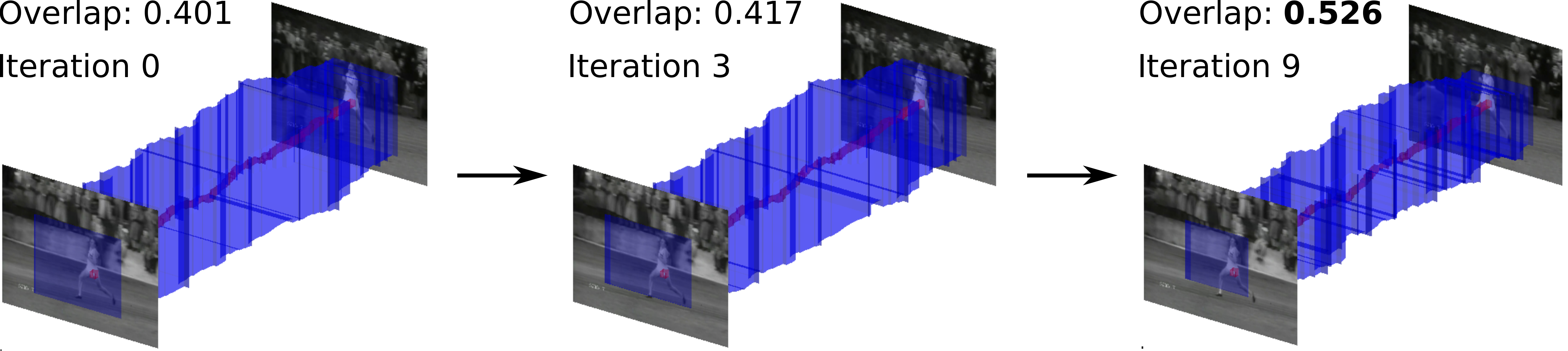}
\caption{\emph{Running.}}
\end{subfigure}
\vspace{0.05cm}\\
\begin{subfigure}{0.875\textwidth}
\centering
\includegraphics[width=\textwidth]{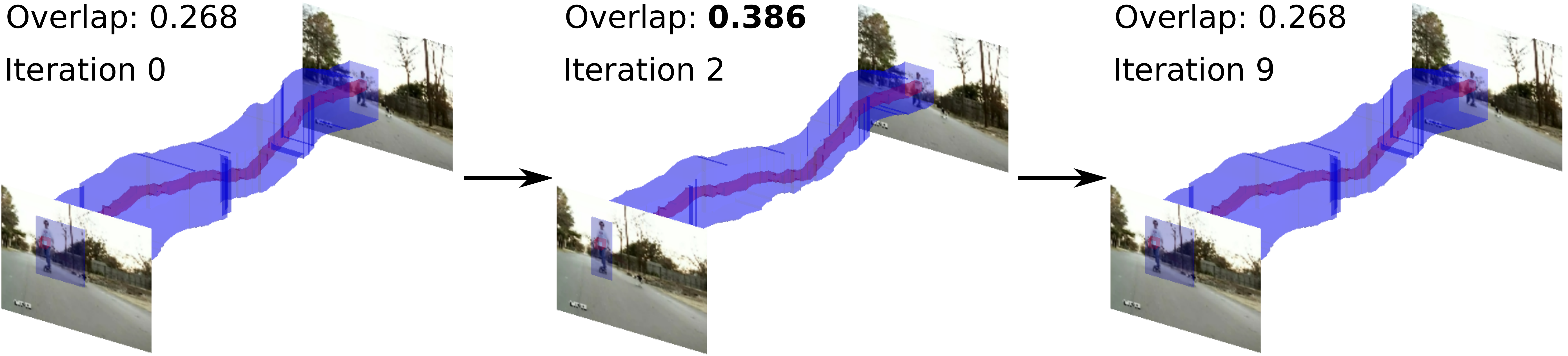}
\caption{\emph{Skateboarding.}}
\end{subfigure}
\caption{\textbf{Qualitative examples} of the iterative proposal mining (blue) during training, guided by points (red) on UCF Sports. (a) and (b): the final best proposals have a significantly improved overlap (from 0.194 to 0.627 and from 0.401 to 0.526 IoU). (c): the final best proposal is the same as the initial best proposal, although halfway through the iterations, a better proposal was mined.}
\label{fig:exp2-qual}
\end{figure}

The results of Figure~\ref{fig:exp2} show the effectiveness of our proposal mining after the iterative optimization. In Figure~\ref{fig:exp2-qual}, we provide three qualitative training examples, highlighting the mining during the iterations. We show two successful examples, where mining improves the quality of the top proposal, and a failure case, where the proposal mining reverts back to the initially mined proposal. 

Based on this experiment, we conclude that points are faster to annotate, while they retain localization performance. We recommend that at least 10\% of the frames are annotated with a point to mine the best proposals during training. Doing so results in a 45 times or more annotation time speed-up.

\subsection{Hollywood2Tubes: Action localization for Hollywood2}
\begin{figure}[t]
\centering
\begin{subfigure}{0.49\textwidth}
\centering
\includegraphics[width=\textwidth]{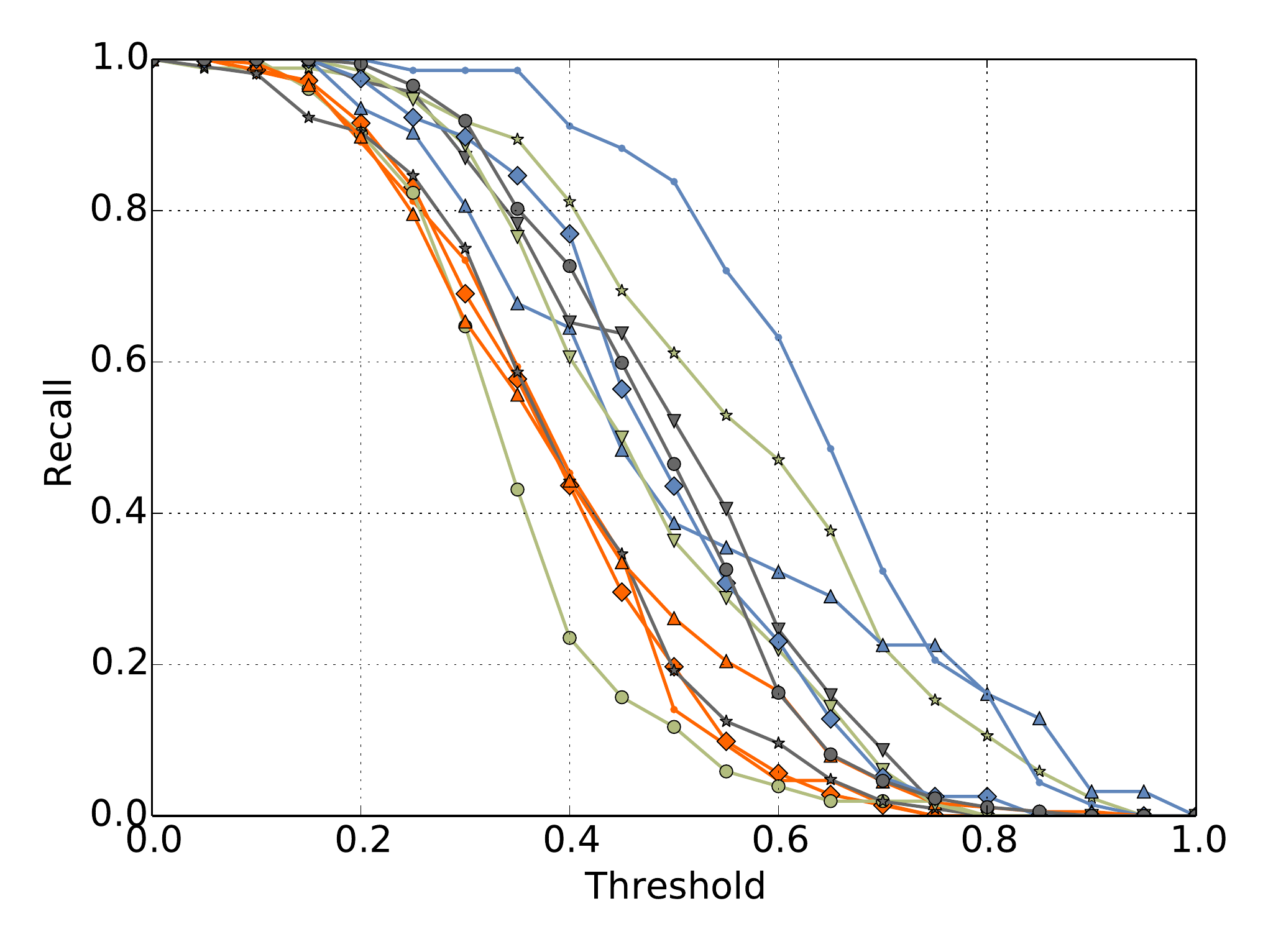}
\caption{Recalls (MABO: 0.47).}
\label{fig:hollywood2-recall}
\end{subfigure}
\begin{subfigure}{0.49\textwidth}
\centering
\includegraphics[width=\textwidth]{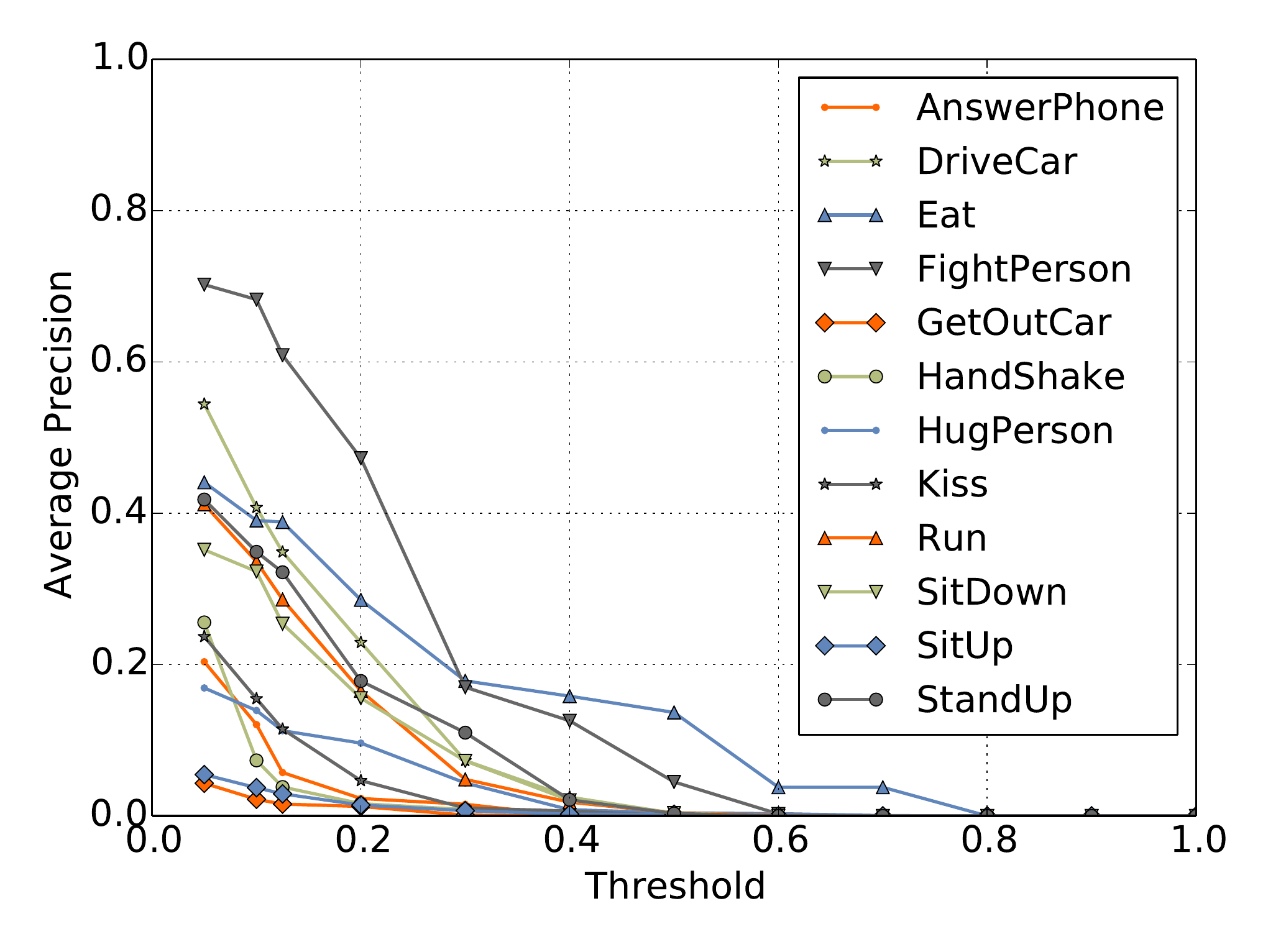}
\caption{Average Precisions.}
\label{fig:hollywood2-map}
\end{subfigure}
\caption{\textbf{Hollywood2Tubes}: Localization results for Hollywood2 actions across all overlap thresholds.
The discrepancy between the recall and Average Precision indicates the complexity of the \emph{Hollywood2Tubes} dataset for action localization.}
\label{fig:hollywood2-res}
\end{figure}

Based on the results from the first two experiments, we are able to supplement the complete Hollywood2 dataset by Marsza{\l}ek \etal \cite{marszalek09} with action location annotations, resulting in \emph{Hollywood2Tubes}. The dataset consists of 12 actions, such as \emph{Answer a Phone}, \emph{Driving a Car}, and \emph{Sitting up/down}. In total, there are 823 train videos and 884 test videos, where each video contains at least one action. Each video can furthermore have multiple instances of the same action. Following the results of Experiment 2 we have annotated a point on each action instance for every 10 frames per training video. In total, there are 1,026 action instances in the training set; 29,802 frames have been considered and 16,411 points have been annotated. For the test videos, we are still required to annotate bounding boxes to perform the evaluation. We annotate every 10 frames with a bounding box. On both UCF Sports and UCF 101, using 1 in 10 frames yields practically the same IoU score on the proposals. In total, 31,295 frames have been considered, resulting in 15,835 annotated boxes. The annotations, proposals, and localization results are available at \texttt{\url{http://tinyurl.com/hollywood2tubes}}.
\\\\
\textbf{Results.} Following the experiments on UCF Sports and UCF 101, we apply proposals~\cite{vangemert2015apt} on the videos of the Hollywood2 dataset. In Figure~\ref{fig:hollywood2-recall}, we report the action localization test recalls based on our annotation efforts. Overall, a MABO of 0.47 is achieved.
The recall scores are lowest for actions with a small temporal span, such as \emph{Shaking hands} and \emph{Answer a Phone}. The recall scores are highest for actions such as \emph{Hugging a person} and \emph{Driving a Car}. This is primarily because these actions almost completely fill the frames in the videos and have a long temporal span.

In Figure~\ref{fig:hollywood2-map}, we show the Average Precision scores using our proposal mining with point overlap scores. We observe that a high recall for an action does not necessarily yield a high Average Precision score. For example, the action \emph{Sitting up} yields an above average recall curve, but yields the second lowest Average Precision curve. The reverse holds for the action \emph{Fighting a Person}, which is a top performer in Average Precision. These results provide insight into the complexity of jointly recognizing and localizing the individual actions of \emph{Hollywood2Tubes}. The results of Figure~\ref{fig:hollywood2-res} shows that there is a lot of room for improvement.

\begin{figure}[t]
\centering
\begin{subfigure}{0.3\textwidth}
\centering
\includegraphics[width=\textwidth]{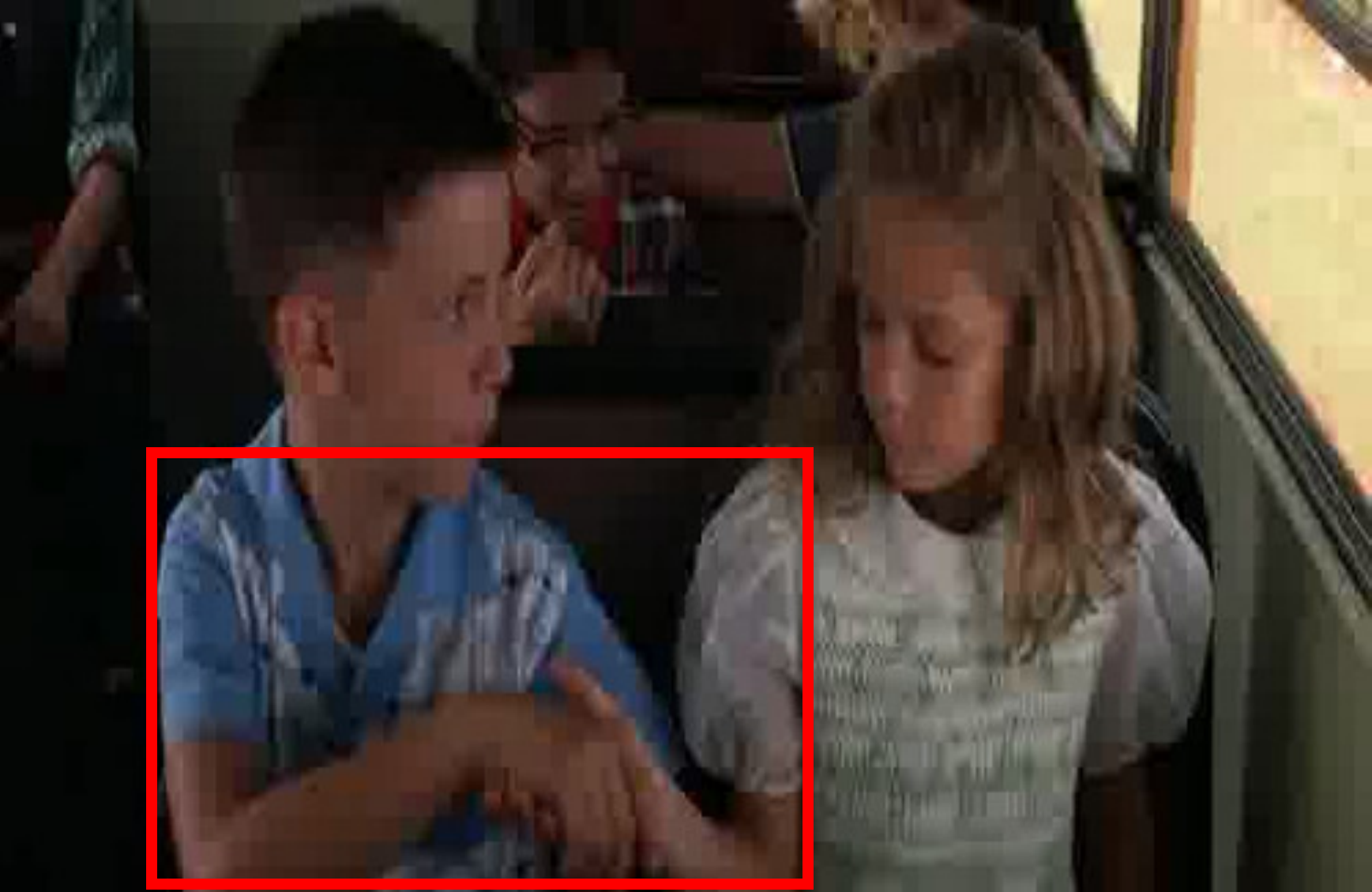}
\caption{Interactions.}
\label{fig:hardcase-1}
\end{subfigure}
\begin{subfigure}{0.3\textwidth}
\centering
\includegraphics[width=\textwidth]{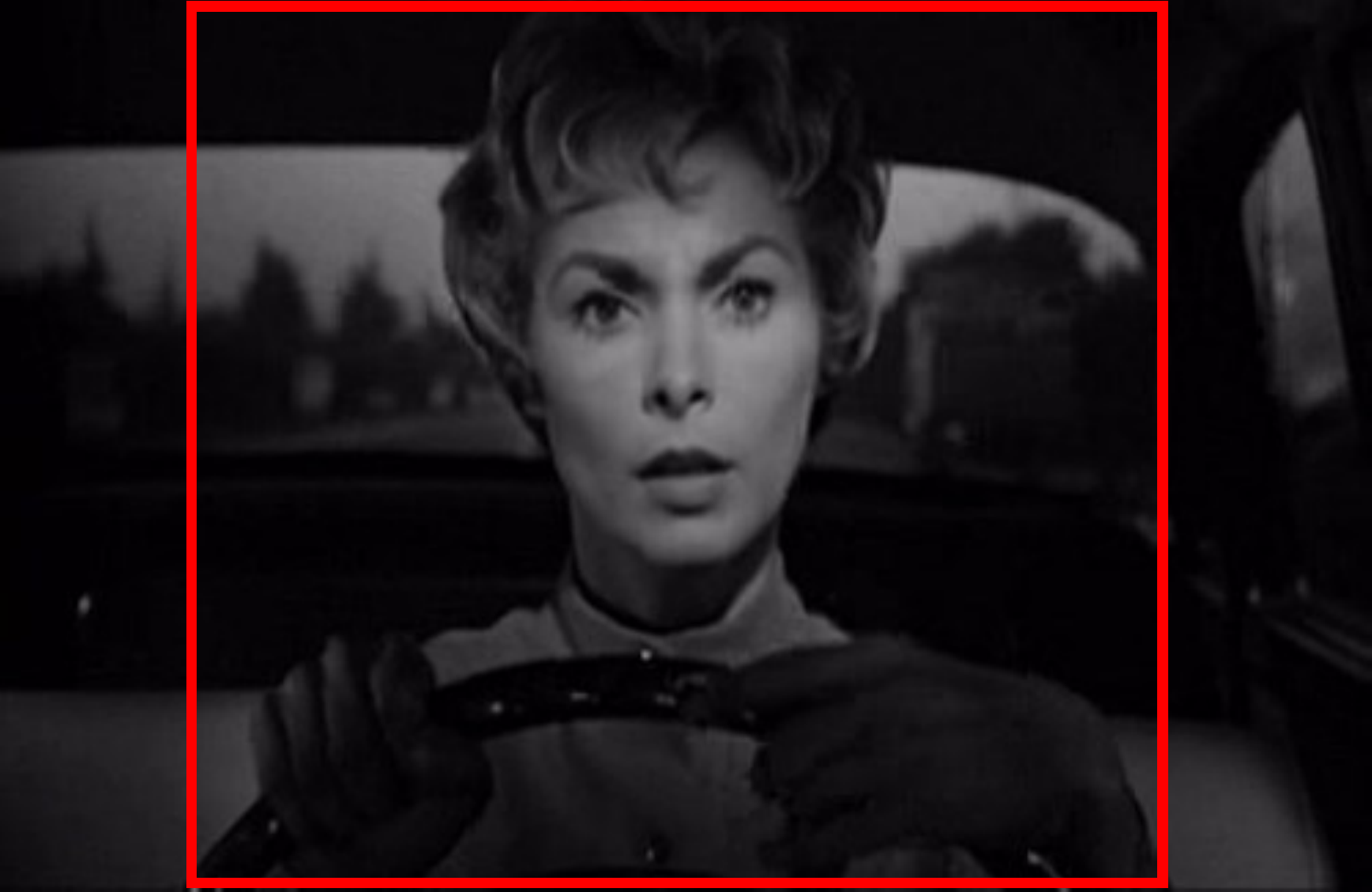}
\caption{Context.}
\label{fig:hardcase-2}
\end{subfigure}
\begin{subfigure}{0.3\textwidth}
\centering
\includegraphics[width=\textwidth]{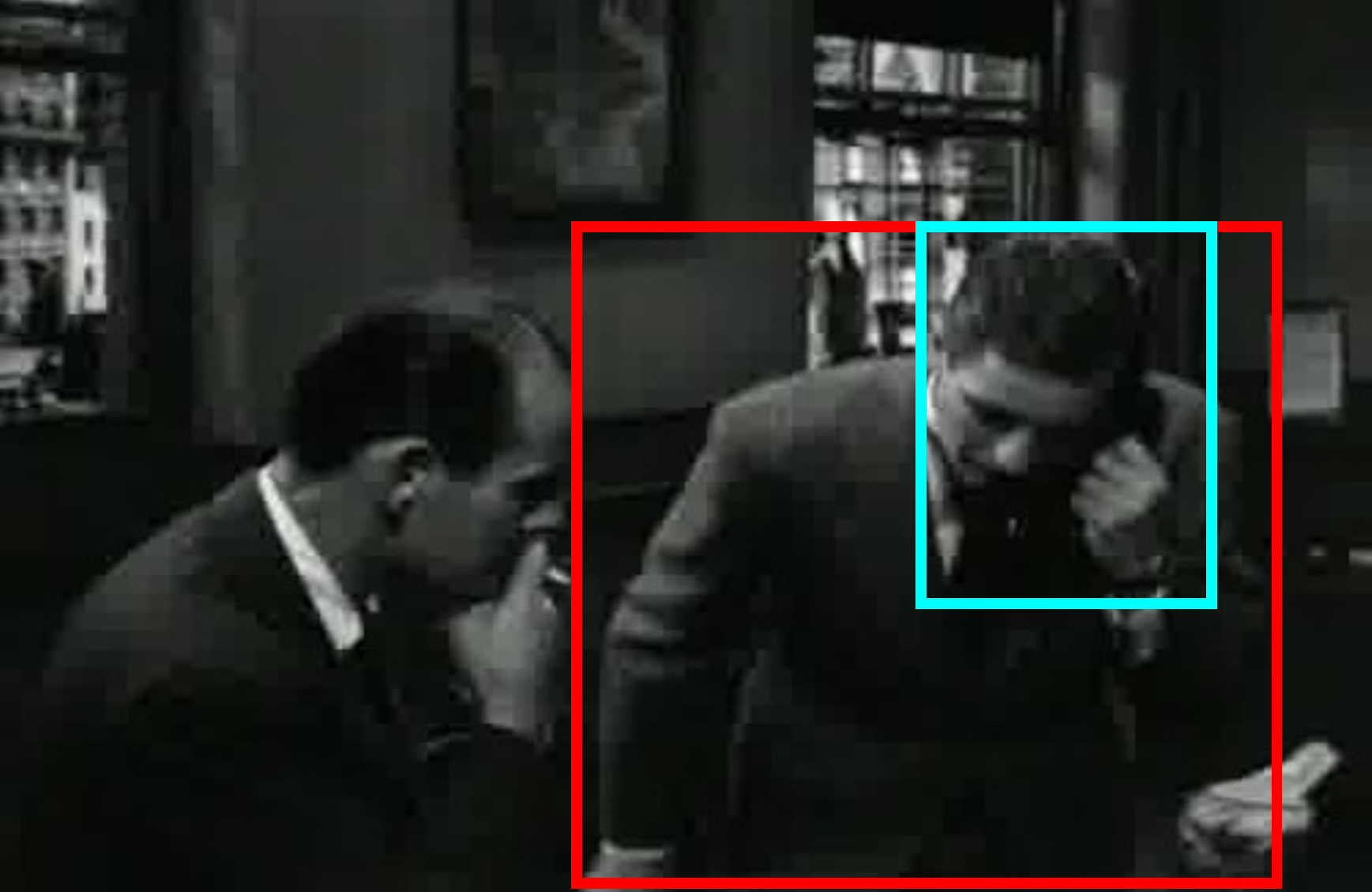}
\caption{Co-occurrence.}
\label{fig:hardcase-3}
\end{subfigure}
\caption{\textbf{Hard scenarios for action localization} using Hollywood2Tubes, not present in current localization challenges. Highlighted are actions involving two or more people, actions partially defined by context, and co-occurring actions within the same video.}
\label{fig:hardcases}
\end{figure}

In Figure~\ref{fig:hardcases}, we highlight a difficult cases for action localization, which are not present in current localization datasets, adding to the complexity of the dataset. In the Supplementary Materials, we outline additional difficult cases, such as cinematographic effects and switching between cameras within the same scene.

\subsection{Comparison to the state-of-the-art}

In the fourth experiment, we compare our results using the point annotations to the current state-of-the-art on action localization using box annotations on the UCF Sports, UCF 101, and Hollywood2Tubes datasets. 
In Table~\ref{tab:sota}, we provide a comparison to related work on all datasets.
For the UCF 101 and Hollywood2Tubes datasets, we report results with the mean Average Precision. For UCF Sports, we report results with the Area Under the Curve (AUC) score, as the AUC score is the most used evaluation score on the dataset. All reported scores are for an overlap threshold of 0.2.

We furthermore compare our results to two baselines using other forms of cheap annotations. This first baseline is the method of Jain \etal~\cite{jain2015objects2action} which performs zero-shot localization, \ie no annotation of the action itself is used, only annotations from other actions. The second baseline is the approach of Cinbis \etal~\cite{cinbis2014multi} using global labels, applied to actions.

\textbf{UCF Sports.} For UCF Sports, we observe that our AUC score is competitive to the current state-of-the-art using full box supervision. Our AUC score of 0.545 is, similar to Experiments 1 and 2, nearly identical to the APT score (0.546)~\cite{vangemert2015apt}. The score is furthermore close to the current state-of-the-art score of 0.559~\cite{gkioxari2015finding,weinzaepfelICCV2015learningToTrack}. The AUC scores for the two baselines without box supervision can not compete with our AUC scores. This result shows that points provide a rich enough source of annotations that are exploited by our proposal mining.

\textbf{UCF 101.} For UCF 101, we again observe similar performance to APT~\cite{vangemert2015apt} and an improvement over the baseline annotation method. The method of Weinzaepfel \etal~\cite{weinzaepfelICCV2015learningToTrack} performs better on this dataset. We attribute this to their strong proposals, which are not unsupervised and require additional annotations.

\textbf{Hollywood2Tubes.} For Hollywood2Tubes, we note that approaches using full box supervision can not be applied, due to the lack of box annotations on the training videos. We can still perform our approach and the baseline method of Cinbis \etal~\cite{cinbis2014multi}. First, observe that the mean Average Precision scores on this dataset are lower than on UCF Sports and UCF 101, highlighting the complexity of the dataset. Second, we observe that the baseline approach using global video labels is outperformed by our approach using points, indicating that points provide a richer source of information for proposal mining than the baselines.

From this experiment, we conclude that our proposal mining using point annotations provides a profitable trade-off between annotation effort and performance for action localization.

\begin{table}[t]
\centering
\scalebox{0.95}{
\begin{tabular}{llrrr}
\toprule
\textbf{Method} & \textbf{Supervision} & \hspace{0.05cm} \textbf{UCF Sports} &  \hspace{0.05cm} \textbf{UCF 101} &  \hspace{0.05cm} \textbf{Hollywood2Tubes}\\
 & & AUC & mAP & mAP\\
\hline
Lan \etal~\cite{lan2011discriminative} & box & 0.380 & - & -\\
Tian \etal~\cite{TianPartCVPR2013} & box & 0.420 & - & -\\
Wang \etal~\cite{wang2014video} & box & 0.470 & - & -\\
Jain \etal~\cite{jain2014action} & box & 0.489 & - & -\\
Chen \etal~\cite{chencorsoICCV2015actiondetectionMotionClustering} & box & 0.528 & - & -\\
van Gemert \etal~\cite{vangemert2015apt} & box & 0.546 & 0.345 & -\\
Soomro \etal~\cite{soomroICCV2015actionLocContextWalk} & box & 0.550 & - & -\\
Gkioxari \etal~\cite{gkioxari2015finding} & box & 0.559 & - & -\\
Weinzaepfel \etal~\cite{weinzaepfelICCV2015learningToTrack} & box & 0.559 & 0.468 & -\\
\hline
Jain \etal~\cite{jain2015objects2action} & zero-shot & 0.232 & - & -\\
Cinbis \etal~\cite{cinbis2014multi}$^{\star}$ & video label & 0.278 & 0.136 & 0.009\\
This work & points & 0.545 & 0.348 & 0.143\\
\bottomrule
\end{tabular}}
\caption{\textbf{State-of-the-art localization results} on the UCF Sports, UCF 101, and Hollywood2Tubes for an overlap threshold of 0.2. Where $^{\star}$ indicates we run the approach of Cinbis \etal~\cite{cinbis2014multi} intended for images on videos. Our approach using point annotations provides a profitable trade-off between annotation effort and performance for action localization.}
\label{tab:sota}
\end{table}

\section{Conclusions}
\label{sec:conclusions}

We conclude that carefully annotated bounding boxes precisely around an action are not needed for action localization. Instead of training on examples defined by expensive bounding box annotations on every frame, we use proposals for training yielding similar results. To determine which proposals are most suitable for training we only require cheap point annotations on the action for a fraction of the frames.
Experimental evaluation on the UCF Sports and UCF 101 datasets shows that:
(i) the use of proposals over directly using the ground truth does not lead to a loss in localization performance,
(ii) action localization using points is comparable to using full box supervision, while being significantly faster to annotate,
(iii) our results are competitive to the current state-of-the-art.
Based on our approach and experimental results we furthermore introduce \emph{Hollywood2Tubes}, a new action localization dataset with point annotations for train videos. The point of this paper is that valuable annotation time is better spent on clicking in more videos than on drawing precise bounding boxes.

\section*{Acknowledgements}
This research is supported by the STW STORY project.

\bibliographystyle{splncs}
\bibliography{1512}

\begin{thebibliography}{10}

\bibitem{TianPartCVPR2013}
Tian, Y., Sukthankar, R., Shah, M.:
\newblock Spatiotemporal deformable part models for action detection.
\newblock In: CVPR. (2013)

\bibitem{jain2014action}
Jain, M., Van~Gemert, J., J{\'e}gou, H., Bouthemy, P., Snoek, C.G.M.:
\newblock Action localization with tubelets from motion.
\newblock In: CVPR. (2014)

\bibitem{yuCVPR2015fap}
Yu, G., Yuan, J.:
\newblock Fast action proposals for human action detection and search.
\newblock In: CVPR. (2015)

\bibitem{vangemert2015apt}
van Gemert, J.C., Jain, M., Gati, E., Snoek, C.G.M.:
\newblock Apt: Action localization proposals from dense trajectories.
\newblock In: BMVC. (2015)

\bibitem{soomroICCV2015actionLocContextWalk}
Soomro, K., Idrees, H., Shah, M.:
\newblock Action localization in videos through context walk.
\newblock In: ICCV. (2015)

\bibitem{KimNIPS2009}
Kim, G., Torralba, A.:
\newblock Unsupervised detection of regions of interest using iterative link
  analysis.
\newblock In: NIPS. (2009)

\bibitem{RussakovskyECCV2012}
Russakovsky, O., Lin, Y., Yu, K., Fei-Fei, L.:
\newblock Object-centric spatial pooling for image classification.
\newblock In: ECCV. (2012)

\bibitem{cinbis2014multi}
Cinbis, R.G., Verbeek, J., Schmid, C.:
\newblock Multi-fold mil training for weakly supervised object localization.
\newblock In: CVPR. (2014)

\bibitem{NguyenICCV2009}
Nguyen, M., Torresani, L., de~la Torre, F., Rother, C.:
\newblock Weakly supervised discriminative localization and classification: a
  joint learning process.
\newblock In: ICCV. (2009)

\bibitem{andrews2002support}
Andrews, S., Tsochantaridis, I., Hofmann, T.:
\newblock Support vector machines for multiple-instance learning.
\newblock In: NIPS. (2002)

\bibitem{XuCVPR2015}
Xu, J., Schwing, A.G., Urtasun, R.:
\newblock Learning to segment under various forms of weak supervision.
\newblock In: CVPR. (2015)

\bibitem{bearmanArXiv15whatsthepoint}
Bearman, A., Russakovsky, O., Ferrari, V., Fei-Fei, L.:
\newblock What's the point: Semantic segmentation with point supervision.
\newblock ECCV (2016)

\bibitem{marszalek09}
Marsza{\l}ek, M., Laptev, I., Schmid, C.:
\newblock Actions in context.
\newblock In: CVPR. (2009)

\bibitem{lan2011discriminative}
Lan, T., Wang, Y., Mori, G.:
\newblock Discriminative figure-centric models for joint action localization
  and recognition.
\newblock In: ICCV. (2011)

\bibitem{gkioxari2015finding}
Gkioxari, G., Malik, J.:
\newblock Finding action tubes.
\newblock In: CVPR. (2015)

\bibitem{weinzaepfelICCV2015learningToTrack}
Weinzaepfel, P., Harchaoui, Z., Schmid, C.:
\newblock Learning to track for spatio-temporal action localization.
\newblock In: ICCV. (2015)

\bibitem{lucorsoCVPR2015humanAction}
Lu, J., Xu, R., Corso, J.J.:
\newblock Human action segmentation with hierarchical supervoxel consistency.
\newblock In: CVPR. (2015)

\bibitem{wang2014video}
Wang, L., Qiao, Y., Tang, X.:
\newblock Video action detection with relational dynamic-poselets.
\newblock In: ECCV.
\newblock (2014)

\bibitem{oneata2014spatio}
Oneata, D., Revaud, J., Verbeek, J., Schmid, C.:
\newblock Spatio-temporal object detection proposals.
\newblock In: ECCV. (2014)

\bibitem{chencorsoICCV2015actiondetectionMotionClustering}
Chen, W., Corso, J.J.:
\newblock Action detection by implicit intentional motion clustering.
\newblock In: ICCV. (2015)

\bibitem{marianICCV2015unsupervisedTube}
Marian~Puscas, M., Sangineto, E., Culibrk, D., Sebe, N.:
\newblock Unsupervised tube extraction using transductive learning and dense
  trajectories.
\newblock In: ICCV. (2015)

\bibitem{soomro2014actionInSports}
Soomro, K., Zamir, A.R.:
\newblock Action recognition in realistic sports videos.
\newblock In: Computer Vision in Sports. (2014)

\bibitem{raptis2012discovering}
Raptis, M., Kokkinos, I., Soatto, S.:
\newblock Discovering discriminative action parts from mid-level video
  representations.
\newblock In: CVPR. (2012)

\bibitem{cao2010crossDatasetActionDetectionMSRIIset}
Cao, L., Liu, Z., Huang, T.S.:
\newblock Cross-dataset action detection.
\newblock In: CVPR. (2010)

\bibitem{soomro2012ucf101}
Soomro, K., Zamir, A.R., Shah, M.:
\newblock Ucf101: A dataset of 101 human actions classes from videos in the
  wild.
\newblock arXiv:1212.0402 (2012)

\bibitem{zhangICCV13actemes}
Zhang, W., Zhu, M., Derpanis, K.:
\newblock From actemes to action: A strongly-supervised representation for
  detailed action understanding.
\newblock In: ICCV. (2013)

\bibitem{jhuangICCV2013towardsUnderstanding}
Jhuang, H., Gall, J., Zuffi, S., Schmid, C., Black, M.:
\newblock Towards understanding action recognition.
\newblock In: ICCV. (2013)

\bibitem{gorban2015thumos}
Gorban, A., Idrees, H., Jiang, Y., Zamir, A.R., Laptev, I., Shah, M.,
  Sukthankar, R.:
\newblock Thumos challenge: Action recognition with a large number of classes.
\newblock In: CVPR workshop. (2015)

\bibitem{karpathy2014largescalevidSports1M}
Karpathy, A., Toderici, G., Shetty, S., Leung, T., Sukthankar, R., Fei-Fei, L.:
\newblock Large-scale video classification with convolutional neural networks.
\newblock In: CVPR. (2014)

\bibitem{kuehne2011hmdb}
Kuehne, H., Jhuang, H., Garrote, E., Poggio, T., Serre, T.:
\newblock Hmdb: a large video database for human motion recognition.
\newblock In: ICCV. (2011)

\bibitem{mihalcik2003ViPER}
Mihalcik, D., Doermann, D.:
\newblock The design and implementation of viper.
\newblock Technical report (2003)

\bibitem{vondrickIJCV2013crowdsourced}
Vondrick, C., Patterson, D., Ramanan, D.:
\newblock Efficiently scaling up crowdsourced video annotation.
\newblock IJCV \textbf{101}(1) (2013)  184--204

\bibitem{yuenICCV09labelmeVideo}
Yuen, J., Russell, B., Liu, C., Torralba, A.:
\newblock Labelme video: Building a video database with human annotations.
\newblock In: ICCV. (2009)

\bibitem{settles2010active}
Settles, B.:
\newblock Active learning literature survey.
\newblock University of Wisconsin, Madison \textbf{52}(55-66) (2010)

\bibitem{vondrick2011video}
Vondrick, C., Ramanan, D.:
\newblock Video annotation and tracking with active learning.
\newblock NIPS (2011)

\bibitem{biancoCVIU15interactiveAnnotation}
Bianco, S., Ciocca, G., Napoletano, P., Schettini, R.:
\newblock An interactive tool for manual, semi-automatic and automatic video
  annotation.
\newblock CVIU \textbf{131} (2015)  88--99

\bibitem{bilenCVPR15weakObjDetConvexClust}
Bilen, H., Pedersoli, M., Tuytelaars, T.:
\newblock Weakly supervised object detection with convex clustering.
\newblock In: CVPR. (2015)

\bibitem{OquabCVPR15isObjLocForFree}
Oquab, M., Bottou, L., Laptev, I., Sivic, J.:
\newblock Is object localization for free? - weakly-supervised learning with
  convolutional neural networks.
\newblock In: CVPR. (2015)

\bibitem{choCVPR15unsupervised}
Cho, M., Kwak, S., Schmid, C., Ponce, J.:
\newblock Unsupervised object discovery and localization in the wild:
  Part-based matching with bottom-up region proposals.
\newblock In: CVPR. (2015)

\bibitem{aliCVPR11flowboost}
Ali, K., Hasler, D., Fleuret, F.:
\newblock Flowboost - appearance learning from sparsely annotated video.
\newblock In: CVPR. (2011)

\bibitem{MisraCVPR15semiSupObjeDetfromVid}
Misra, I., Shrivastava, A., Hebert, M.:
\newblock Watch and learn: Semi-supervised learning for object detectors from
  video.
\newblock In: CVPR. (2015)

\bibitem{wangECCV14videoObject}
Wang, L., Hua, G., Sukthankar, R., Xue, J., Zheng, N.:
\newblock Video object discovery and co-segmentation with extremely weak
  supervision.
\newblock In: ECCV. (2014)

\bibitem{sivaECCV12defenceNegativeMining}
Siva, P., Russell, C., Xiang, T.:
\newblock In defence of negative mining for annotating weakly labelled data.
\newblock In: ECCV. (2012)

\bibitem{kwakICCV15unsupervisedObjectInVid}
Kwak, S., Cho, M., Laptev, I., Ponce, J., Schmid, C.:
\newblock Unsupervised object discovery and tracking in video collections.
\newblock In: ICCV. (2015)

\bibitem{mosabbeb2014multi}
Mosabbeb, E.A., Cabral, R., De~la Torre, F., Fathy, M.:
\newblock Multi-label discriminative weakly-supervised human activity
  recognition and localization.
\newblock In: ACCV.
\newblock (2014)

\bibitem{siva2011weakly}
Siva, P., Xiang, T.:
\newblock Weakly supervised action detection.
\newblock In: BMVC. (2011)

\bibitem{jain2015objects2action}
Jain, M., van Gemert, J.C., Mensink, T., Snoek, C.G.M.:
\newblock Objects2action: Classifying and localizing actions without any video
  example.
\newblock In: ICCV. (2015)

\bibitem{tseng2009quantifying}
Tseng, P.H., Carmi, R., Cameron, I.G., Munoz, D.P., Itti, L.:
\newblock Quantifying center bias of observers in free viewing of dynamic
  natural scenes.
\newblock JoV \textbf{9}(7) (2009)

\bibitem{RodriguezCVPR2008}
Rodriguez, M.D., Ahmed, J., Shah, M.:
\newblock Action {MACH}: a spatio-temporal maximum average correlation height
  filter for action recognition.
\newblock In: CVPR. (2008)

\bibitem{wang13}
Wang, H., Schmid, C.:
\newblock Action recognition with improved trajectories.
\newblock In: ICCV. (2013)

\bibitem{sanchez2013image}
S{\'a}nchez, J., Perronnin, F., Mensink, T., Verbeek, J.:
\newblock Image classification with the fisher vector: Theory and practice.
\newblock IJCV \textbf{105}(3) (2013)  222--245

\end{thebibliography}

\clearpage
\section*{Supplementary materials}

The supplementary materials for the ECCV paper "Spot On: Action Localization from Pointly-Supervised Proposals" contain the following elements regarding \emph{Hollywood2Tubes}:
\begin{itemize}
\item The annotation protocol for the dataset.
\item Annotation statistics for the train and test sets.
\item Visualization of box annotations for each action.
\end{itemize}

\section*{Annotation protocol}
Below, we outline how each action is specifically annotated using a bounding box. The protocol is the same for the point annotations, but only the center of the box is annotated, rather than the complete box.
\begin{itemize}
\item \textbf{AnswerPhone:} A box is drawn around both the head of the person answering the phone and the hand holding the phone (including the phone itself), from the moment the phone is picked up.
\item \textbf{DriveCar:} A box is drawn around the person in the driver seat, including the upper part of the stear itself. In case of a video clip with of a driving car in the distance, rather than a close-up of the people in the car, the whole car is annotated as the driver can hardly be distinguished.
\item \textbf{Eat:} A single box is drawn around the union of the people who are joinly eating.
\item \textbf{FightPerson:} A box is drawn around both people fighting for the duration of the fight. If only a single person is visible, no annotation is made. In case of a chaotic brawl with more than two people, a single box is drawn around the union of the fight.
\item \textbf{GotOutCar:} A box is drawn around the person starting from the moment that the first body parts exists the car until the person is standing complete outside the car, beyond the car door.
\item \textbf{HandShake:} A box is drawn around the complete arms (the area between the union of the shoulders, ellbows, and hands) of the people shaking hands.
\item \textbf{HugPerson:} A box is drawn around the heads and upper torso (until the waist, if visible) of both hugging people.
\item \textbf{Kiss:} A box is drawn around the heads of both kissing people.
\item \textbf{Run:} A box is drawn around the running person.
\item \textbf{SitDown:} A box is drawn around the complete person from the moment the person starts moving down until the person is complete seated at rest.
\item \textbf{SitUp:} A box is drawn around the complete person from the moment the person starts to move upwards from a laid down position until the person no longer moves upwards..
\item \textbf{StandUp:} Vice versa to SitDown.
\end{itemize}

\begin{figure}[t]
\centering
\begin{subfigure}{0.45\textwidth}
\includegraphics[width=\textwidth]{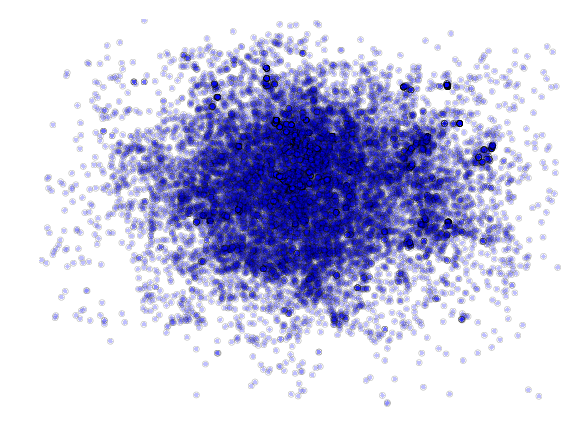}
\caption{Points (train).}
\end{subfigure}
\begin{subfigure}{0.45\textwidth}
\includegraphics[width=\textwidth]{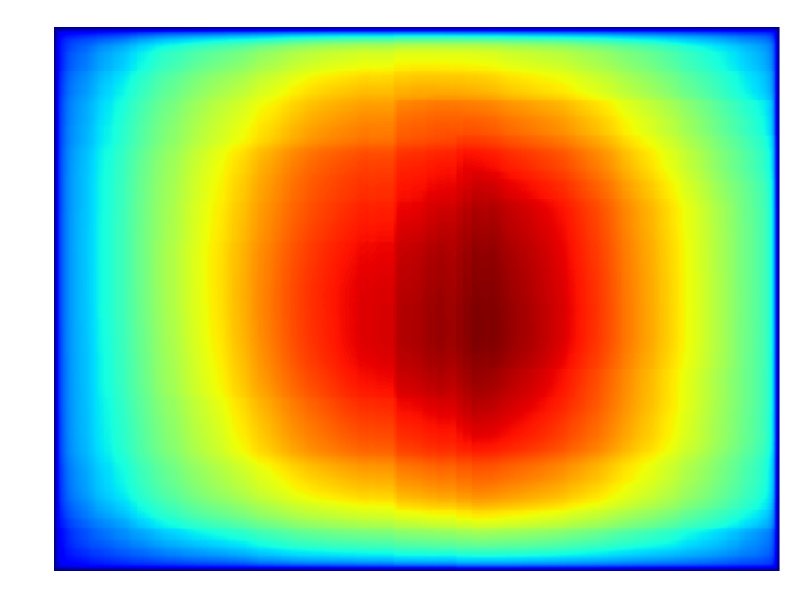}
\caption{Boxes (test).}
\end{subfigure}
\caption{Annotation aggregations for the point and box annotations on \emph{Hollywood2Tubes}. The annotations are overall center-oriented, but we do note a bias towards the rule-of-thirds principle, given the higher number of annotations on $\frac{2}{3}$-th the width of the frame.}
\label{fig:stats}
\end{figure}

\begin{table}[t]
\centering
\begin{tabular}{l r r}
\toprule
 & \hspace{1cm} Training set & \hspace{1cm} Test set\\
\midrule
Number of videos & 823 & 884\\
Number of action instances & 1,026 & 1,086\\
Numbers of frames evaluated & 29,802 & 31,295\\
Number of annotations & 16,411 & 15,835\\
\bottomrule
\end{tabular}
\caption{Annotation statistics for \emph{Hollywood2Tubes}. The large difference between the number of frames evaluated and the number of annotations is because the actions in Hollwood2 are not trimmed.}
\label{tab:stats-all}
\end{table}

\section*{Annotation statistics}
In Figure~\ref{fig:stats}, we show the aggregated point annotations (training set) and box annotations (test set). The aggregation shows that the localization is center oriented. The heatmap for the box annotations do show the rule-of-thirds principle, given the the higher number of annotations on $\frac{2}{3}$-th the width of the frame.

In Table~\ref{tab:stats-all}, we show a number of statistics on the annotations performed on the dataset.

\section*{Annotation examples}
\begin{figure}[h]
\centering
\begin{subfigure}{0.3\textwidth}
\includegraphics[width=\textwidth]{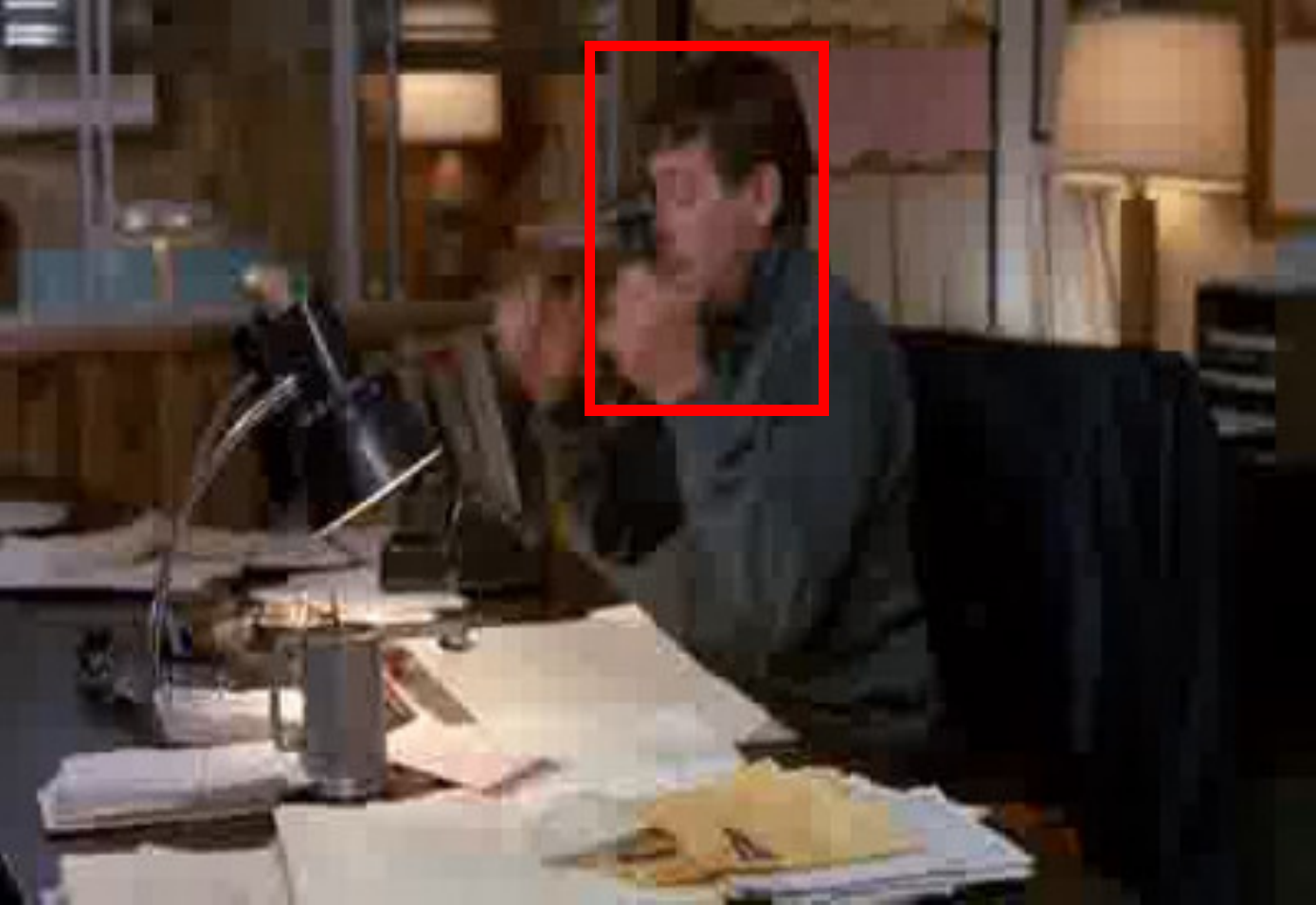}
\caption{Answer Phone.}
\end{subfigure}
\begin{subfigure}{0.3\textwidth}
\includegraphics[width=\textwidth]{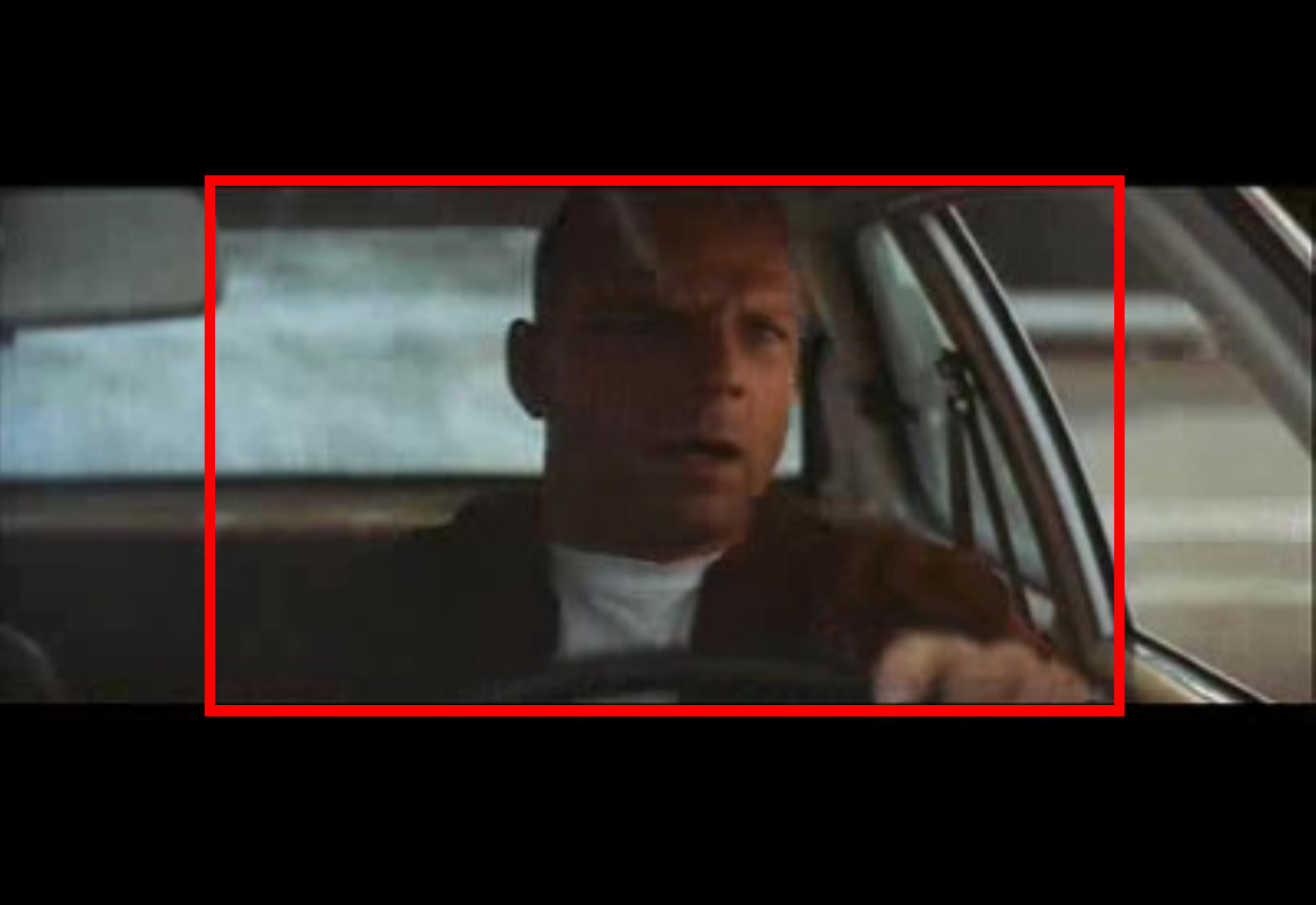}
\caption{Drive Car.}
\end{subfigure}
\begin{subfigure}{0.3\textwidth}
\includegraphics[width=\textwidth]{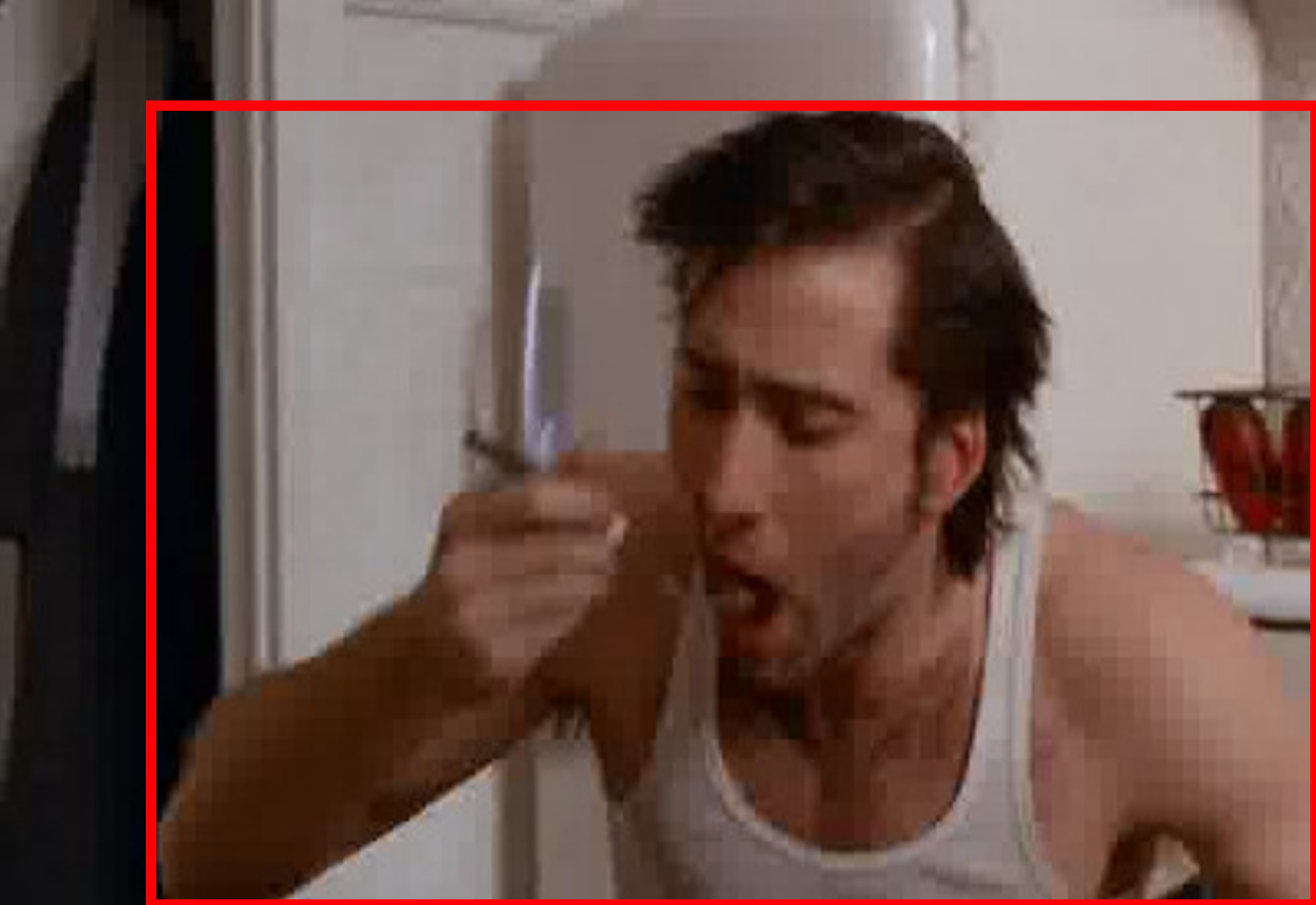}
\caption{Eat.}
\end{subfigure}
\begin{subfigure}{0.3\textwidth}
\includegraphics[width=\textwidth]{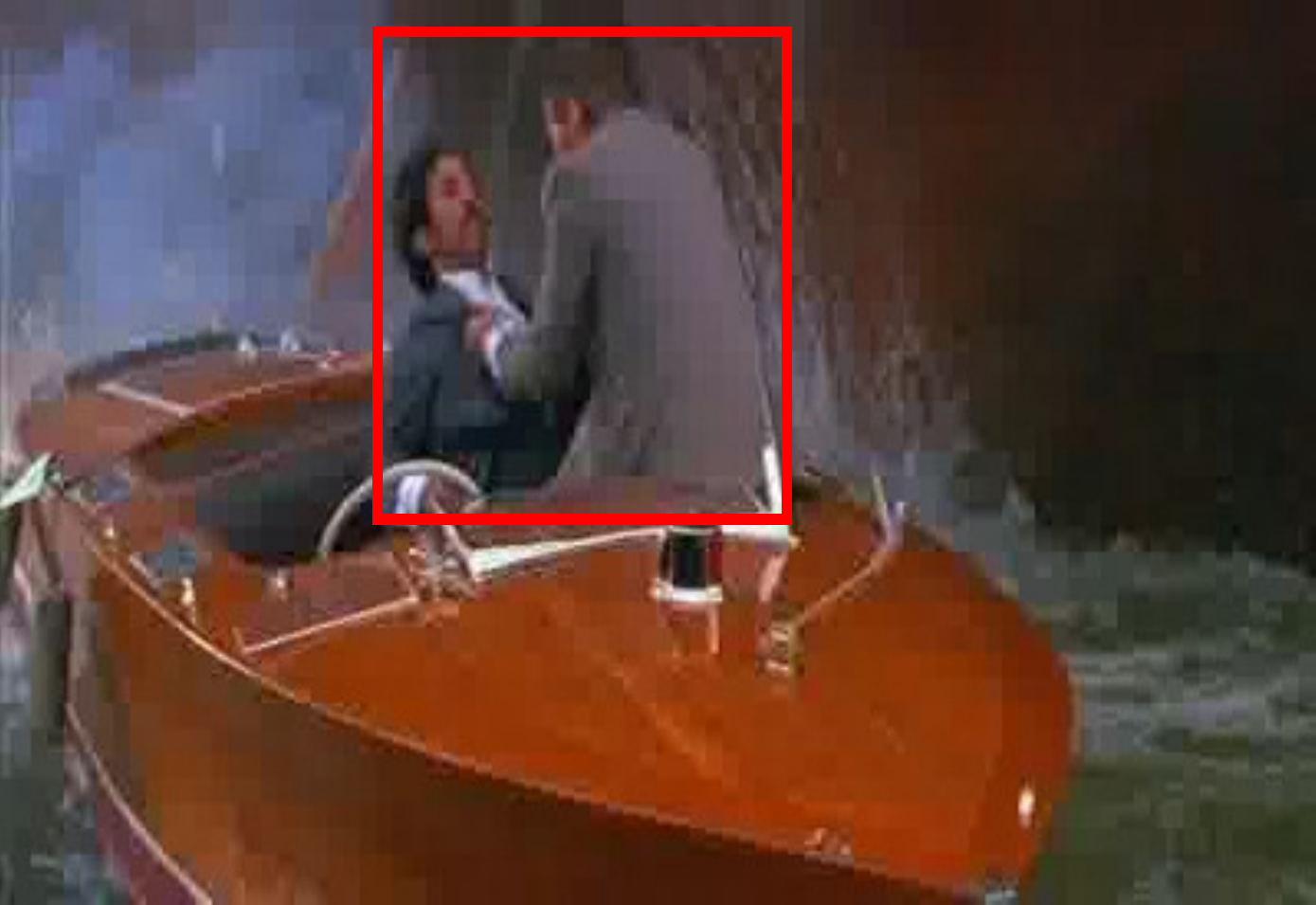}
\caption{Fight Person.}
\end{subfigure}
\begin{subfigure}{0.3\textwidth}
\includegraphics[width=\textwidth]{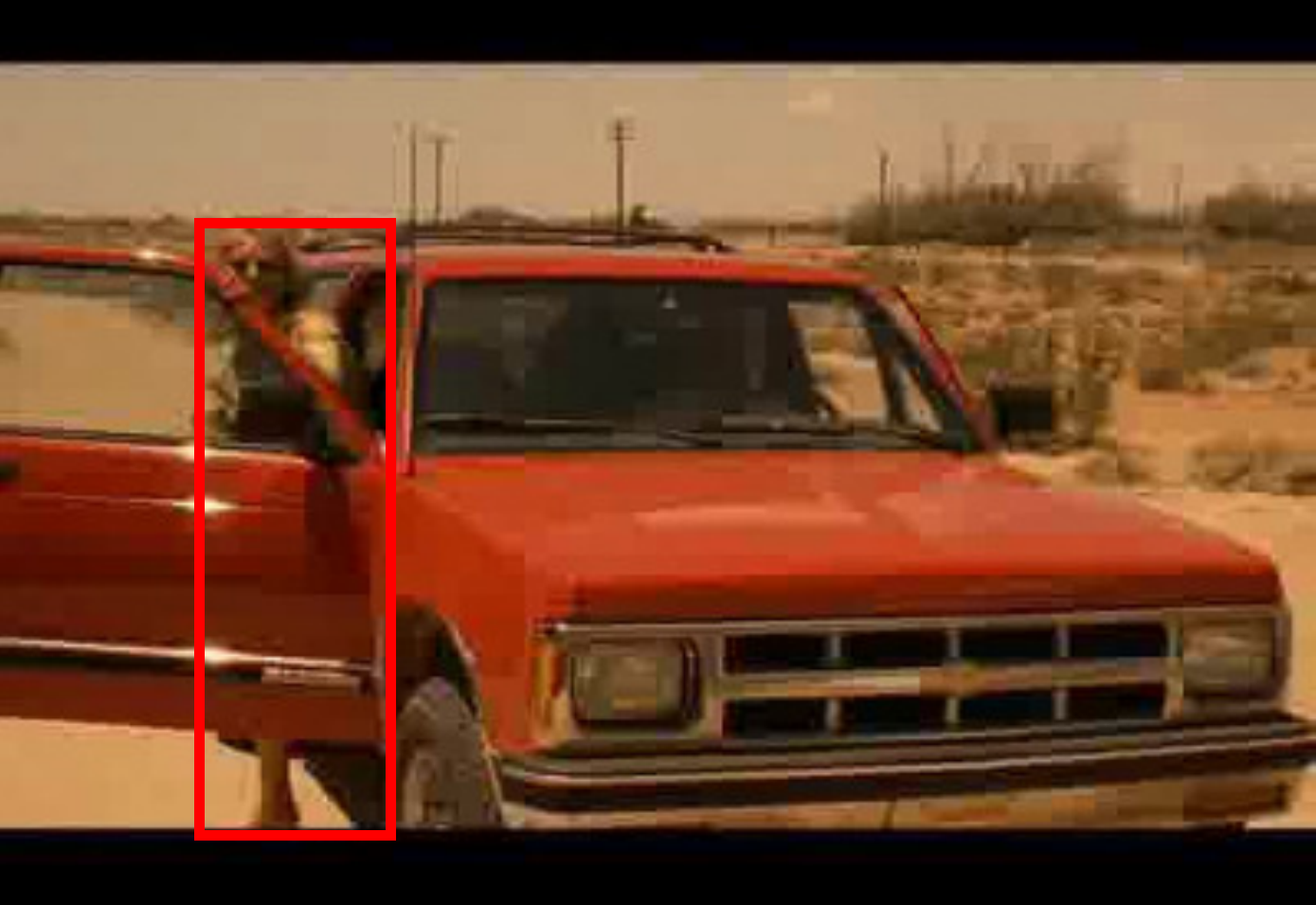}
\caption{Get out of Car.}
\end{subfigure}
\begin{subfigure}{0.3\textwidth}
\includegraphics[width=\textwidth]{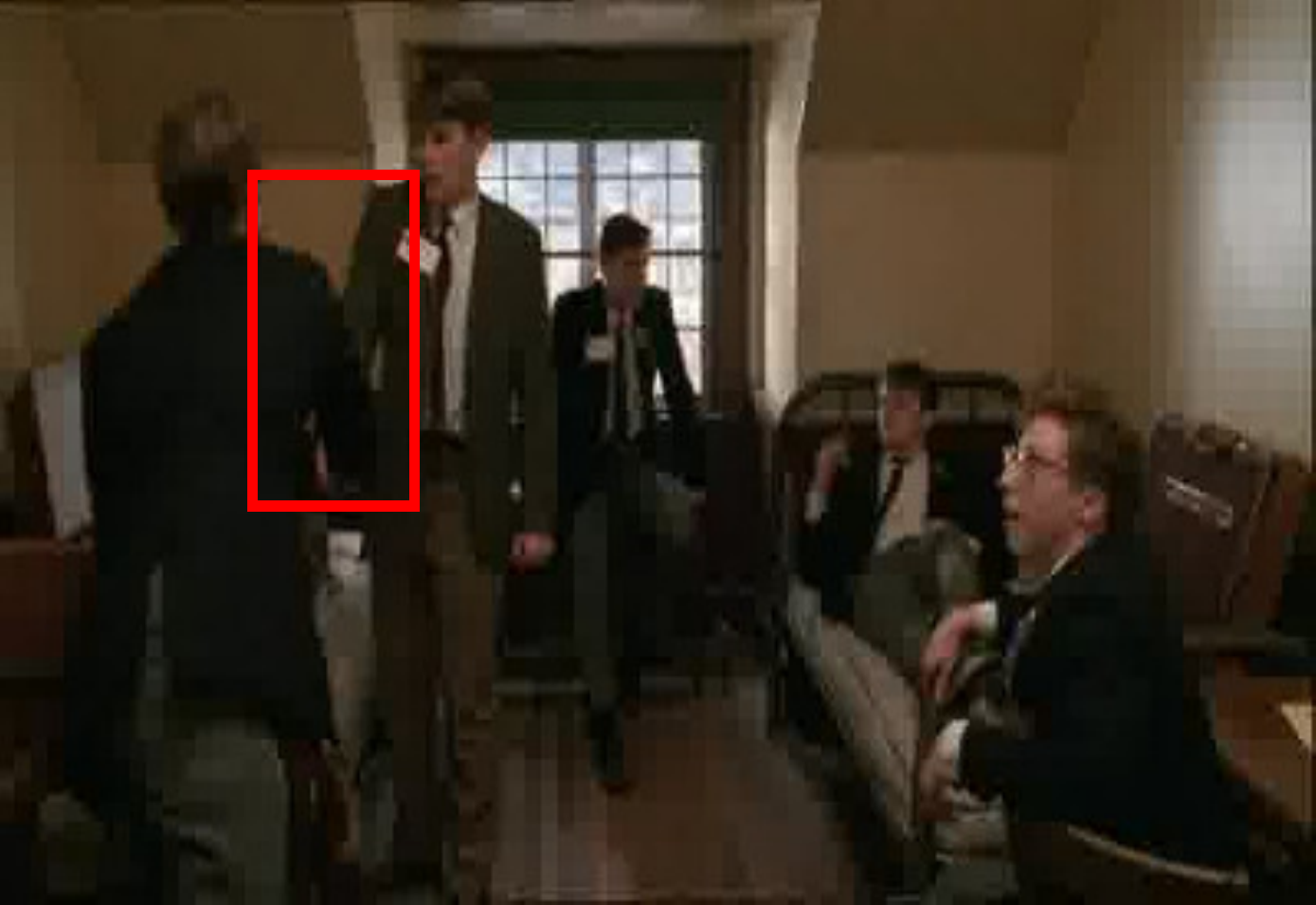}
\caption{Hand Shake.}
\end{subfigure}
\begin{subfigure}{0.3\textwidth}
\includegraphics[width=\textwidth]{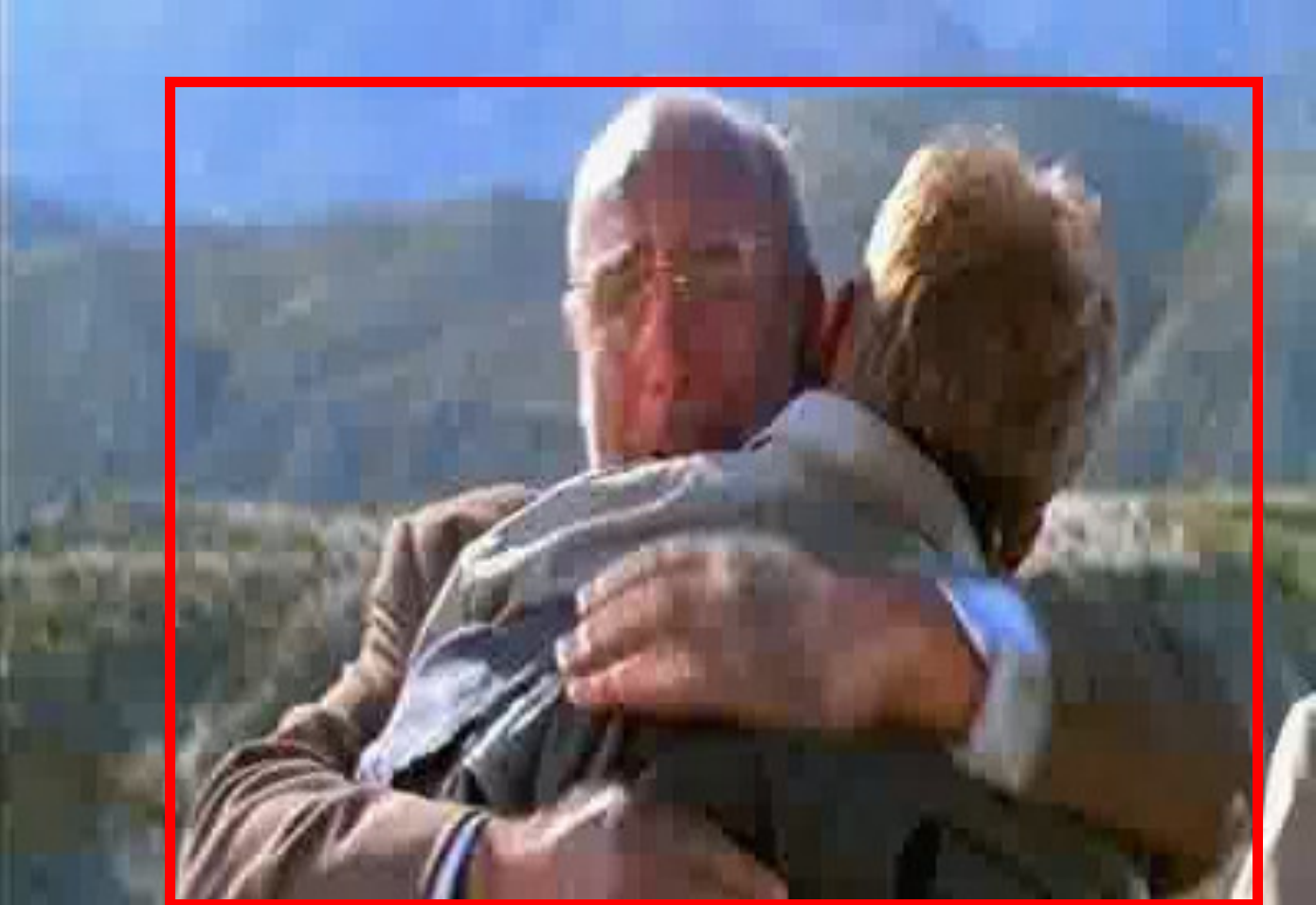}
\caption{Hug.}
\end{subfigure}
\begin{subfigure}{0.3\textwidth}
\includegraphics[width=\textwidth]{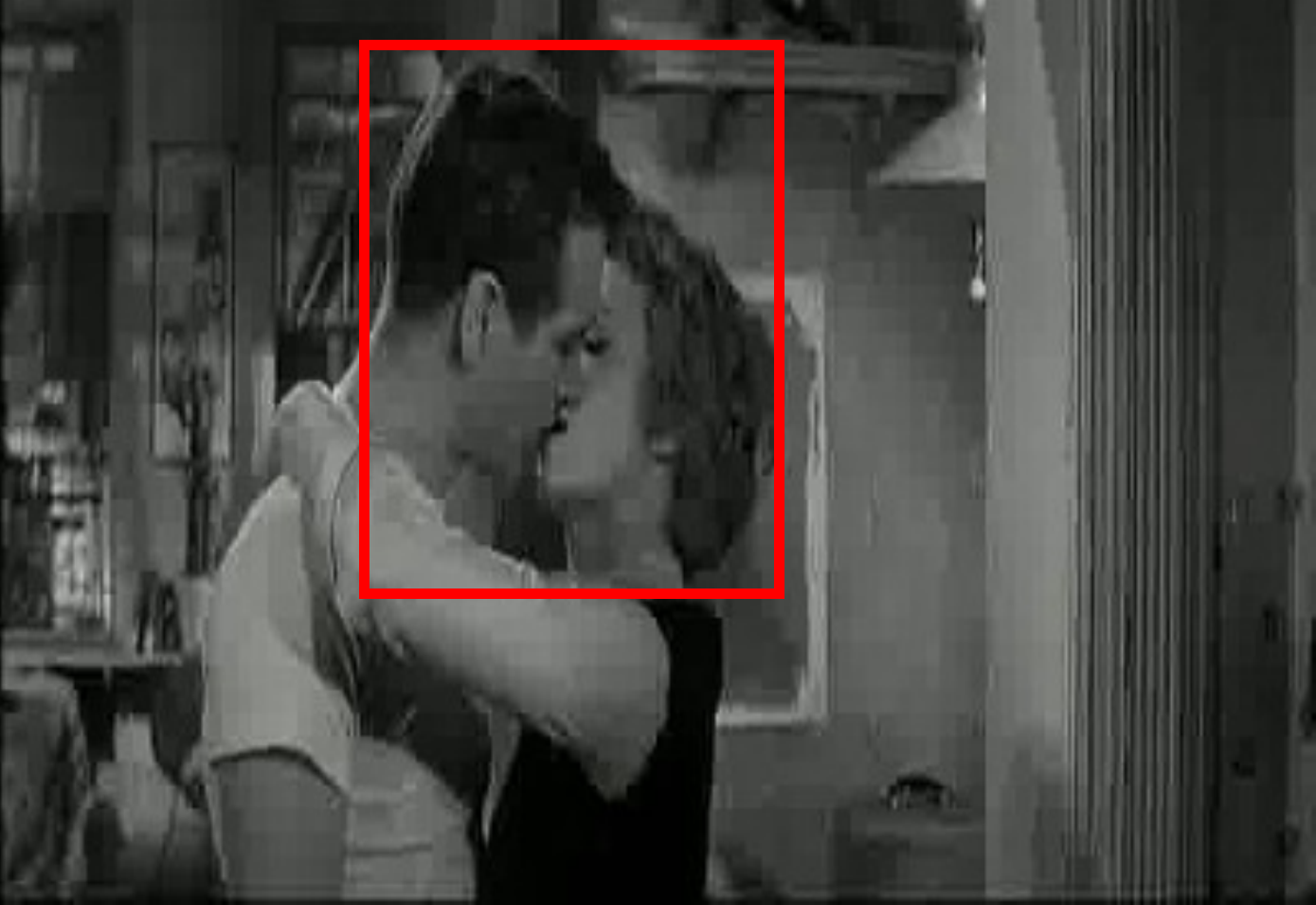}
\caption{Kiss.}
\end{subfigure}
\begin{subfigure}{0.3\textwidth}
\includegraphics[width=\textwidth]{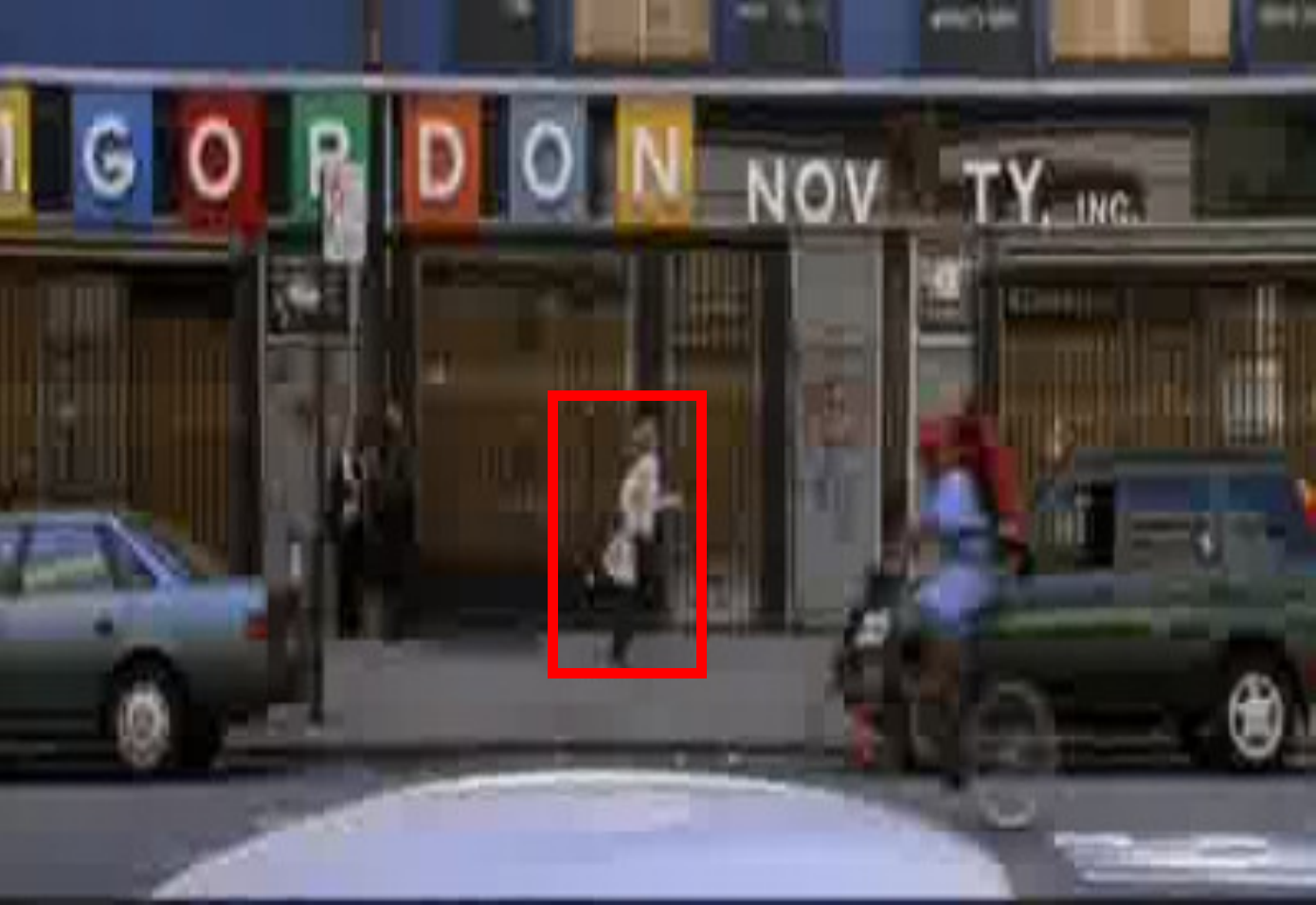}
\caption{Run.}
\end{subfigure}
\begin{subfigure}{0.3\textwidth}
\includegraphics[width=\textwidth]{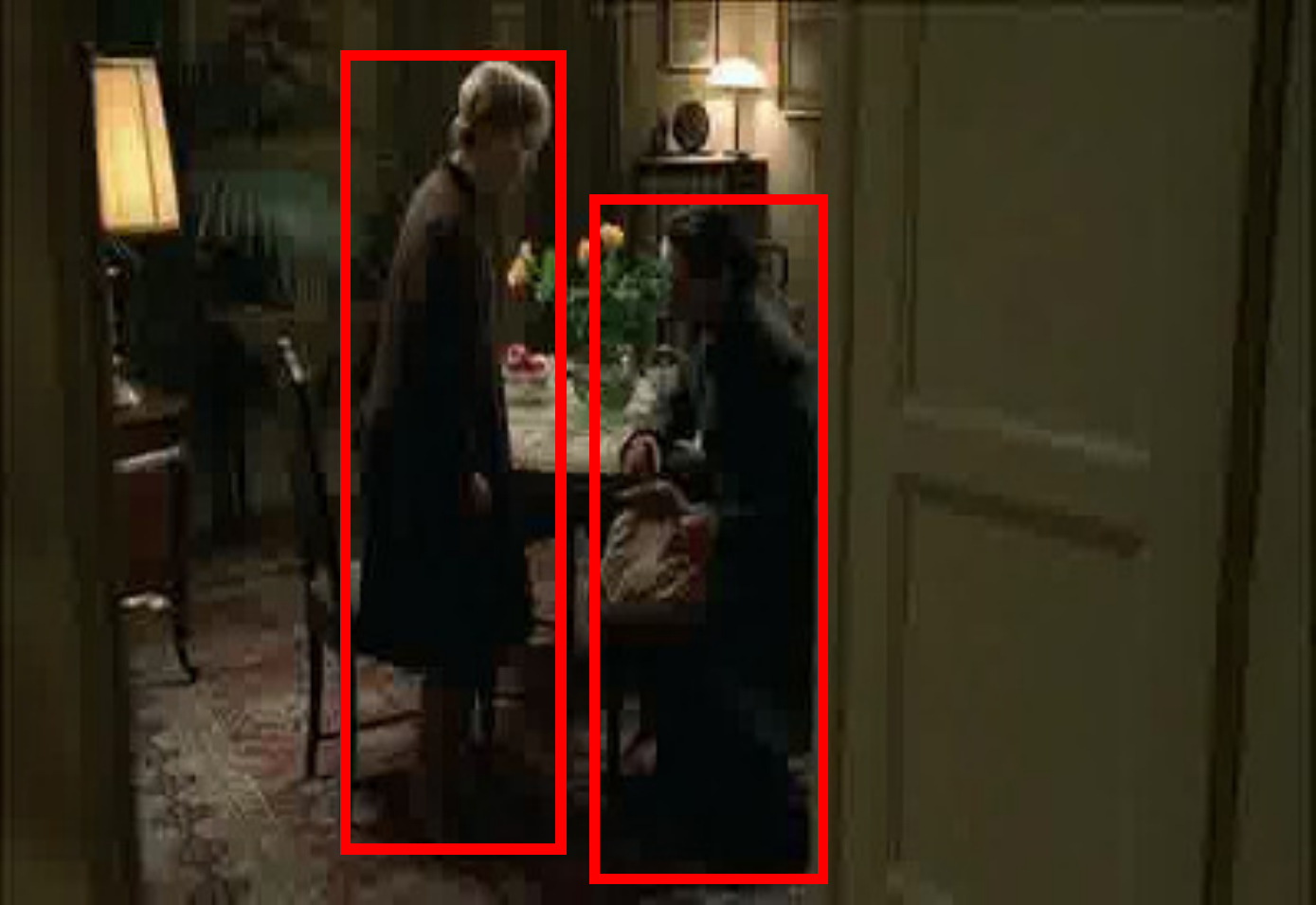}
\caption{Sit down.}
\end{subfigure}
\begin{subfigure}{0.3\textwidth}
\includegraphics[width=\textwidth]{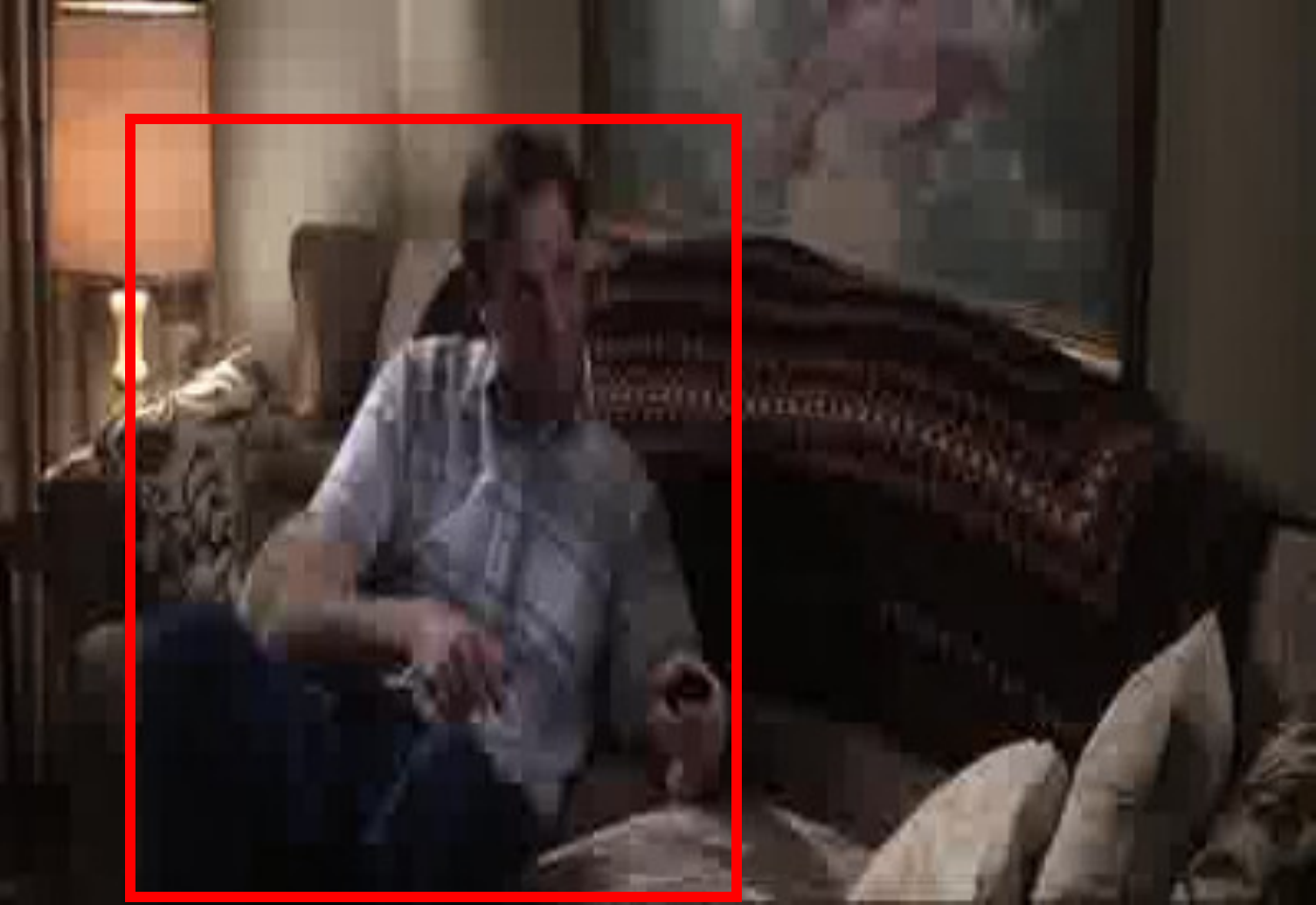}
\caption{Sit up.}
\end{subfigure}
\begin{subfigure}{0.3\textwidth}
\includegraphics[width=\textwidth]{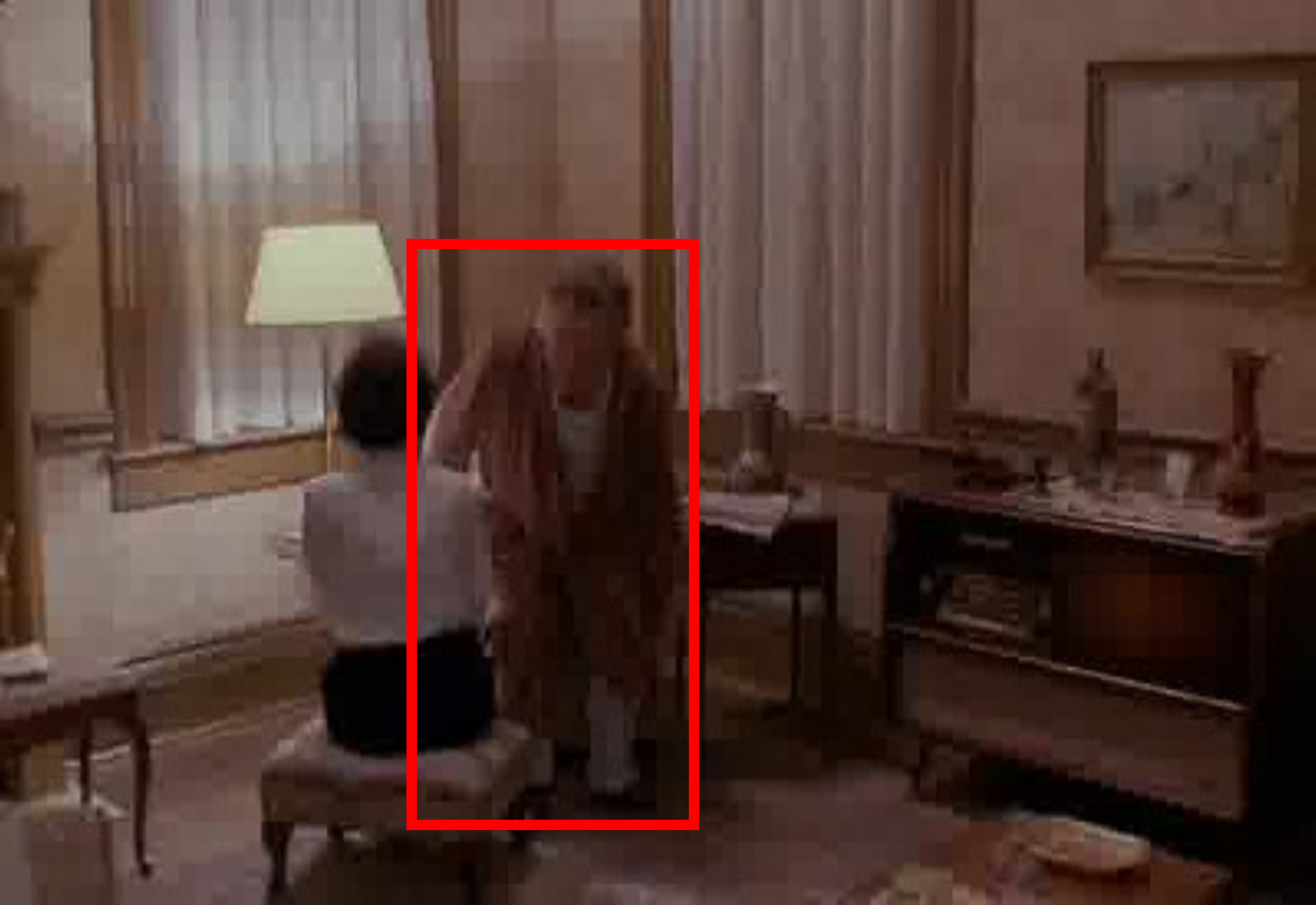}
\caption{Stand up.}
\end{subfigure}
\caption{Example box annotations of test videos for \emph{Hollywood2Tubes}.}
\label{fig:h2t-examples}
\end{figure}

In Figure~\ref{fig:h2t-examples} we show an example frame of each of the 12 actions, showing the diversity and complexity of the videos for action localization.

\end{document}